\documentclass[pdflatex,sn-vancouver,Numbered]{sn-jnl}

\usepackage{graphicx}%
\usepackage{rotating}
\usepackage{adjustbox}
\usepackage{multirow}%
\usepackage{upgreek}
\usepackage{amsmath,amssymb,amsfonts}%
\usepackage{amsthm}%
\usepackage{mathrsfs}%
\usepackage[title]{appendix}%
\usepackage{xcolor}%
\usepackage{textcomp}%
\usepackage{manyfoot}%
\usepackage{booktabs}%
\usepackage{algorithm}%
\usepackage{algorithmicx}%
\usepackage{algpseudocode}%
\usepackage{listings}%
\usepackage{subcaption}
\usepackage{url}
\usepackage{tabularx}
\usepackage{array}

\theoremstyle{thmstyleone}%
\theoremstyle{thmstyletwo}%

\theoremstyle{thmstylethree}%

\begin{document}

\title[Article Title]{AutoTandemML: Active Learning Enhanced Tandem Neural Networks for Inverse Design Problems }

\author*[1]{\fnm{Luka} \sur{Grbcic}}\email{lgrbcic@lbl.gov}

\author[2]{\fnm{Juliane} \sur{M\"uller}}\email{juliane.mueller@nrel.gov}

\author*[1]{\fnm{Wibe Albert} \sur{de Jong}}\email{wadejong@lbl.gov}

\affil*[1]{\orgdiv{Applied Mathematics and Computational Research}, \orgname{Lawrence Berkeley National Laboratory}, \orgaddress{\street{1 Cyclotron Rd}, \city{Berkeley}, \postcode{94720}, \state{California}, \country{USA}}}

\affil[2]{\orgdiv{Computational Science Center}, \orgname{National Renewable Energy Laboratory}, \orgaddress{\street{15013 Denver West Parkway}, \city{Golden}, \postcode{80401}, \state{Colorado}, \country{USA}}}

\abstract{Inverse design in science and engineering involves determining optimal design parameters that achieve desired performance outcomes, a process often hindered by the complexity and high dimensionality of design spaces, leading to significant computational costs. To tackle this challenge, we propose a novel hybrid approach that combines active learning with Tandem Neural Networks to enhance the efficiency and effectiveness of solving inverse design problems.  Active learning allows to selectively sample the most informative data points, reducing the required dataset size without compromising accuracy. We investigate this approach using three benchmark problems: airfoil inverse design, photonic surface inverse design, and scalar boundary condition reconstruction in diffusion partial differential equations. We  demonstrate that integrating active learning with Tandem Neural Networks outperforms standard approaches across the benchmark suite, achieving better accuracy with fewer training samples.}


\maketitle

\section{Introduction}\label{sec:intro}

Inverse design problems are inherently complex because they require determining the necessary inputs or configurations to achieve a specific desired outcome, often within systems governed by non-linear, multi-dimensional relationships. These challenges are prevalent across various engineering disciplines—for instance, designing an airfoil shape to meet certain aerodynamic criteria \cite{li2022machine}, or designing photonic materials based on target optical properties \cite{molesky2018inverse, grbcic2025artificial}. Such problems are frequently ill-posed, lacking unique solutions and being sensitive to data imperfections, which worsens their complexity. Consequently, there is a critical need for efficient computational tools and advanced methods—such as optimization algorithms and machine learning—to solve inverse design problems effectively, enabling engineers to innovate and refine designs by working backwards from desired performance specifications.

Deep Neural Networks (DNN) have been very successful as inverse design models in science and engineering \cite{li2022machine, molesky2018inverse, lei2021deep}. Generative Adversarial Networks (GANs) \cite{yilmaz2020conditional, yonekura2022generating}, Variational Autoencoders (VAEs) \cite{tang2020generative, kudyshev2020machine}, and Invertible Neural Networks (INNs) \cite{glaws2022invertible, frising2023tackling} are some of the most prominent DNN-based inverse design approaches. However, both GANs and VAEs have  drawbacks related to training difficulty, instability, and limitations in capturing the full diversity of possible designs \cite{kossale2022mode}, while INNs introduce complex neural architecture components in order to be effective, and thus are harder to train. 

Tandem Neural Networks (TNNs) offer advantages over VAEs, GANs, and INNs in inverse design tasks by enabling direct optimization of input parameters to achieve desired outputs, leading to more stable and straightforward training. They avoid issues like training instability, mode collapse, and difficulty capturing multi-modal outputs that commonly affect VAEs and GANs, resulting in more accurate, and interpretable designs that better meet specific target properties. Furthermore, TNNs do not require the complex neural architectures of INNs, making them more effective and efficient for exploring complex design spaces in inverse design applications.

TNNs have emerged as a powerful tool for solving inverse design problems across various fields, enabling the efficient optimization of complex systems by mapping desired outputs back to input parameters. In the realm of photonics and nanophotonics, they have been extensively employed to design metasurfaces and nanophotonic devices with tailored optical properties. Studies have demonstrated the use of TNNs for optimizing metasurfaces for specific functionalities such as polarization conversion, absorption, and color filtering \cite{noureen2023physics, xu2024inverse, he2023constrained, chen2023inverse, yeung2021designing, yeung2021multiplexed, xie2023deep, liu2023efficient, wang2024inverse}. 

In addition to metasurfaces, TNNs have been applied to the inverse design of other nanophotonic structures, including optical nanoantennas and multilayer thin films \cite{yuan2024bootstrap, swe2024inverse, head2022inverse, xu2021improved, qiu2021simultaneous, liu2024deep}. These studies showcase the versatility of TNNs in handling both continuous and discrete design parameters, facilitating the creation of devices with customized optical responses and advancing the capabilities in nanofabrication and materials science.

Beyond photonics, TNNs have found applications in diverse areas such as porous media transport, radar engineering, wind turbine airfoil design, electronic integrated systems, and optical fiber design. In porous media transport, TNNs have been utilized to model multicomponent reactive transport processes, enhancing the accuracy and efficiency of simulations \cite{chen2021improved}. In radar engineering, they have facilitated the design of composite radar-absorbing structures by enabling inverse design capabilities \cite{nielsen2022design}. Similarly, in wind turbine airfoil design, TNNs have been employed to achieve desired aerodynamic properties by inversely mapping performance metrics to geometric parameters \cite{anand2024novel}. 
In electronic integrated systems, TNNs have assisted in channel inverse design \cite{ma2022channel}. In the field of optical fibers, they have been applied to the inverse design of hollow-core anti-resonant fibers \cite{meng2023artificial}.

Despite the demonstrated efficacy of TNNs in specific applications,  there remains a lack of comprehensive research evaluating their performance across a broad spectrum of inverse design problems. Most existing studies focus on tailored solutions for individual challenges, which limits the understanding of TNNs as a general-purpose tool. In particular, the impact of training data characteristics on the efficacy of TNNs has not been thoroughly investigated. Addressing this gap, the novelty and contributions of our work are as follows:

\begin{enumerate}
	\item We develop a hybrid framework, AutoTandemML, that combines  sampling data generation by active learning with TNNs specifically for inverse design problems. 
	
	\item We investigate whether datasets optimized for accurately predicting forward relationships also perform well in capturing inverse relationships with the TNNs.
	
	\item We compare AutoTandemML with TNNs trained on data generated by other sampling algorithms—namely Random, Latin Hypercube, Best Candidate, and GreedyFP—across three inverse design benchmark problems \cite{kamath2022intelligent}. 
		
	\item We introduce an inverse design benchmark suite comprising three significant problems in science and engineering: Airfoil Inverse Design (AID), Photonic Surfaces Inverse Design (PSID), and Scalar Boundary Reconstruction (SBR) of a scalar diffusion partial differential equation (PDE). We make these benchmarks available to the research community through a public data repository.
	
	\item We develop the AutoTandemML Python module available for everyone—an easy-to-use software tool for the automated generation of TNNs.
\end{enumerate}

The remainder of this article is organized as follows. Sec. \ref{sec:autotandemml} introduces the AutoTandemML framework, including descriptions of inverse design models, and each component that forms the framework—specifically, active learning for multi-output regression and TNNs. Sec. \ref{sec:benchmarks_num_exp} describes the inverse design benchmarks and the setup of all numerical experiments. We outline the sampling methods used for comparison, as well as the accuracy metrics and inverse design validation methods employed in our study. In Sec. \ref{sec:results}, we present the performance results of AutoTandemML and compare it with other sampling methods used to train the TNNs across all inverse design benchmark problems. Finally, App. \ref{app:hyperparams}, \ref{app:benchmarks}, and \ref{app:forward_model} provide additional details on the hyperparameters used for all machine learning models, specifics of the inverse design benchmarks, and supplementary results that enhance the interpretation of AutoTandemML's performance, respectively. This organizational structure aims to guide the reader through the development, implementation, and evaluation of the AutoTandemML framework in a logical and coherent manner.

\section{AutoTandemML Framework}\label{sec:autotandemml}

In this section, we mathematically define inverse design models, the active learning algorithm for multi-output regression and describe the components of the TNN. Furthermore, we explain how active learning and TNNs can be integrated into a single framework for efficiently solving inverse design problems.

\subsection{Inverse Design Models}\label{sec:inversedesign}
 The main objective of inverse design models is to infer design parameters that yield a known target value or desired property. Mathematically, the inverse design problem is defined as:
\begin{equation}
	\mathbf{x} = f^{-1}(\mathbf{y})
	\label{eqn:inverse_design_definition}
\end{equation}

In Equation (\ref{eqn:inverse_design_definition}), $\mathbf{x} \in \mathbb{R}^d$ is the design vector we aim to determine, and $\mathbf{y} \in \mathbb{R}^N$ is the target vector specified \textit{a priori}. The function $f: \mathbb{R}^d \rightarrow \mathbb{R}^N$ represents the forward model that maps design parameters to their resulting properties. In inverse design  we seek to invert this function to find the design parameters that produce the desired target outcomes.

Inverse design problems are typically ill-posed, meaning that multiple values of $\mathbf{x}$ can yield similar values of $\mathbf{y}$. This lack of a unique solution necessitates the use of computational methods and machine learning models to approximate the inverse mapping or to identify suitable design parameters through optimization techniques.

\subsection{Active Learning for Multi-output Regression}

Active learning is a machine learning approach that operates as a form of optimal experimental design, where computational methods are employed to identify and obtain data instances that will most significantly enhance the model's accuracy or performance. By strategically selecting the most informative data points—typically those about which the model is least certain—active learning seeks to maximize learning efficiency (\cite{settles2011theories}). 

More specifically, the active learning algorithm aims to select input design vectors where the machine learning model \( \mathbf{M} \) (also denoted as the forward model) exhibits the highest predictive uncertainty. This is formally expressed as an optimization problem:

\begin{equation}
	\mathbf{x}^*_i \in \underset{\mathbf{x}}{\operatorname{arg\,max}} \, u(\mathbf{x}) \in \underset{\mathbf{x}}{\operatorname{arg\,max}} \sum_{j=1}^{d} \sigma_j(\mathbf{x}; \mathbf{M})
	\label{eq:optimization_problem}
\end{equation}

Here, \( u(\mathbf{x}) \) represents the total uncertainty at input design vector \( \mathbf{x} \in \mathbb{R}^d\), computed as the sum of standard deviations across all \( d \) output dimensions for a multi-output regression problem. $\mathbf{x}^*_i$ represents the $i^{th}$ optimal input design vector based on the maximum value of \( u(\mathbf{x}) \). The standard deviations are encapsulated in the vector:

\begin{equation}
	\boldsymbol{\sigma}(\mathbf{x}_i^*; \mathbf{M}) = \begin{bmatrix}
		\sigma_1(\mathbf{x}_i^*; \mathbf{M}) \\
		\sigma_2(\mathbf{x}_i^*; \mathbf{M}) \\
		\vdots \\
		\sigma_d(\mathbf{x}_i^*; \mathbf{M})
	\end{bmatrix} \in \mathbb{R}^d
	\label{eq:std_dev_vector}
\end{equation}

The algorithm proceeds by iteratively selecting the most uncertain inputs, querying the High fidelity (HF) surrogate \( \mathbf{H} \) at these points, and updating the model accordingly. The detailed steps are outlined in Alg.~\ref{alg:AL}. 

More specifically, the algorithm begins with an initial dataset $\mathcal{D}$ of size $n_0$, where each design vector $\mathbf{x}_i$ is evaluated using the HF surrogate $\mathbf{H}$ to obtain $f(\mathbf{x}_i)$. The model $\mathbf{M}$ is initially trained on this dataset. In each iteration, while the total number of evaluations is less than the pre-defined evaluation budget $N_{max}$, the algorithm performs an optimization step to find $k$ input points $\mathbf{x}^*_i$ that maximize the total predictive uncertainty $u(\mathbf{x}$) of $\mathbf{M}$, as defined in  Eq. \ref{eq:optimization_problem}, each subsequent optimization is started with a new random seed. The $k$ points form the batch $\mathcal{B}$. Each design vector in $\mathcal{B}$ is evaluated using the HF surrogate $\mathbf{H}$ to obtain new outputs, and the dataset $\mathcal{D}$ is updated accordingly. The model $\mathbf{M}$ is retrained on this expanded dataset before the next iteration begins, thereby incrementally improving its performance by focusing on areas where it is most uncertain.

\begin{algorithm}[H]
	\caption{Active Learning for Multi-output Regression}\label{alg:AL}
	\begin{algorithmic}[1]
		\Require
		\begin{itemize}
			\item HF surrogate \( \mathbf{H}(\mathbf{x}) \)
			\item Initial dataset size \( n_0 \)
			\item Maximum number of evaluations \( N_{\text{max}} \)
			\item Batch size \( k \)
		\end{itemize}
		\State Initialize dataset \( \mathcal{D} = \{ ( \mathbf{x}_i, f(\mathbf{x}_i) ) \}_{i=1}^{n_0} \), where \( f(\mathbf{x}_i) = \mathbf{H}(\mathbf{x}_i) \)
		\State Train initial model \( \mathbf{M} \) on dataset \( \mathcal{D} \)
		\While{number of evaluations \( < N_{\text{max}} \)}
		\State \textbf{Optimization Step}:
		\For{ \( i = 1 \) to \( k \) }
		\State Find point \( \mathbf{x}^*_i \) by solving with a new random seed:
		\[
		\mathbf{x}^*_i = \underset{\mathbf{x}}{\operatorname{arg\,max}} \ u(x) = \underset{\mathbf{x}}{\operatorname{arg\,max}} \sum_{j=1}^{d} \sigma_j(\mathbf{x}; \mathbf{M})
		\]
		\EndFor
		\State \( \mathcal{B} = \{ \mathbf{x}^*_{(1)}, \mathbf{x}^*_{(2)}, \dots, \mathbf{x}^*_{(k)} \} \)
        \For{ \( i = 1 \) to \( k \) }
		\State Evaluate HF surrogate: \( f(\mathbf{x}_i) = \mathbf{H}(\mathbf{x}_i) \)
		\State Update dataset: \( \mathcal{D} \leftarrow \mathcal{D} \cup \{ (\mathbf{x}_i, f(\mathbf{x}_i)) \} \)
		\EndFor
		\State Retrain model \( \mathbf{M} \) on updated dataset \( \mathcal{D} \)
		\EndWhile
	\end{algorithmic}
\end{algorithm}

The input design vector ($\mathbf{x}$) and the HF surrogate $\mathbf{H}$ both vary depending on the specific benchmark problem under consideration. The specifics of the design vectors and responses for each benchmark are provided in Sec. \ref{sec:inverse_design_benchmarks}, while detailed descriptions of the HF surrogates $\mathbf{H}$ are given in App. \ref{app:benchmarks}. Details on the computational implementation of the multi-output regression active learning algorithm—including the hyperparameters used, the selection of the model $\mathbf{M}$, and the optimizer employed to determine the optimal design vectors ($\mathbf{x}_i^*$)—can be found in Sec. \ref{sec:num_exp_setup}.

\subsection{Active Learning Enhanced Tandem Neural Networks}

\begin{figure}[H]
	\centering
	\begin{subfigure}[b]{0.495\textwidth}
		\centering
		\includegraphics[height=5cm]{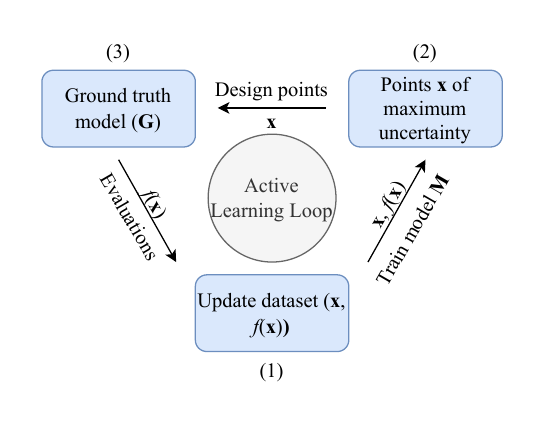}
		\caption{}
		\label{fig:Figure1a}
	\end{subfigure}%

	\begin{subfigure}[b]{0.495\textwidth}
		\centering
		\includegraphics[width=1.0051\linewidth]{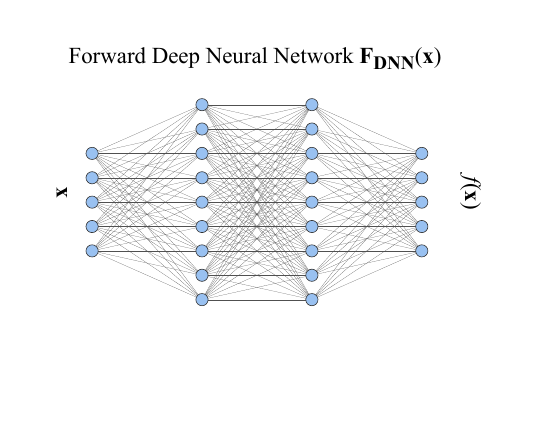}
		\caption{}
		\label{fig:Figure1b}
	\end{subfigure}
	\hfill%
	\begin{subfigure}[b]{0.495\textwidth}
		\centering
		\includegraphics[width=\linewidth]{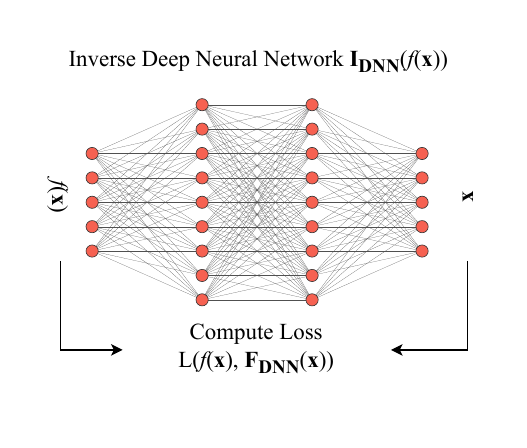}
		\caption{}
		\label{fig:Figure1c}
	\end{subfigure}
	\caption{AutoTandemML framework segments: (a) Active learning to generate a dataset ($\mathbf{x}$, $\mathit{f}(\mathbf{x})$). (b) Forward deep neural network $\mathbf{F_{DNN}}$ training with the active learning generated dataset ($\mathbf{x}$, $\mathit{f}(\mathbf{x})$). (c) Inverse deep neural network $\mathbf{I_{DNN}}$ training with the active learning generated dataset ($\mathit{f}(\mathbf{x})$, $\mathbf{x}$), and a modified loss function that utilizes the $\mathbf{F_{DNN}}$ predictions.}
	\label{fig:Figure1}
\end{figure}

We aim to utilize the dataset $\mathcal{D} = \{ ( \mathbf{x}, f(\mathbf{x}) ) \}$, generated through active learning, to train each component of the TNN. The complete AutoTandemML framework is illustrated in Fig. \ref{fig:Figure1}. Specifically, Fig. \ref{fig:Figure1a} depicts the active learning algorithm. As elaborated in the previous section, the goal of active learning is to employ the model $\mathbf{M}$ to generate design vectors $\mathbf{x}$ that maximize uncertainty, and then evaluate them using the HF surrogate $\mathbf{H}$ to form and update the dataset $\mathcal{D} = \{ ( \mathbf{x}, f(\mathbf{x}) ) \}$.

Furthermore, Figs. \ref{fig:Figure2b} and \ref{fig:Figure2c} illustrate the two segments of the TNN—the forward DNN (denoted as $\mathbf{F_{DNN}}$) and the inverse DNN (denoted as $\mathbf{I_{DNN}}$), respectively. The forward DNN approximates the mapping from input parameters to outputs, while the inverse DNN predicts the input parameters corresponding to desired outputs. By training these networks using the active learning generated dataset, we enhance the predictive accuracy and reliability of the inverse design process.

More specifically, in inverse design, the objective is to determine the input parameters $\mathbf{x}$ that produce a desired output through a complex system. We start by training the  $\mathbf{F_{DNN}}$, to approximate the mapping from inputs to outputs, $\mathbf{x} \rightarrow f(\mathbf{x})$. We then train the $\mathbf{I_{DNN}}$, which aims to predict the input $\mathbf{x} = \mathbf{I_{DNN}}(f(\mathbf{x}))$ corresponding to a given output $f(\mathbf{x})$. During the training of $\mathbf{I_{DNN}}$, we utilize a loss function $L$ that compares the original output $f(\mathbf{x})$ with the output of the forward model when fed with the inverse prediction:
\begin{equation}
	L = L\big(f(\mathbf{x}), \mathbf{F_{DNN}}(\mathbf{x})\big) = L\big(f(\mathbf{x}), \mathbf{F_{DNN}}(\mathbf{I_{DNN}}(f(\mathbf{x})))\big)
\end{equation}
Minimizing this loss adjusts the weights of $\mathbf{I_{DNN}}$ so that the predicted input generates an output through $\mathbf{F_{DNN}}$ that closely matches the desired output $f(\mathbf{x})$.

This tandem training approach ensures that the inverse neural network $\mathbf{I_{DNN}}$ not only seeks inputs that reproduce the desired outputs but also aligns with the learned mapping of the forward neural network $\mathbf{F_{DNN}}$. By incorporating $\mathbf{F_{DNN}}$ into the loss function, we effectively regularize $\mathbf{I_{DNN}}$, promoting reliable inverse solutions. This method is particularly advantageous in scenarios where the inverse problem is ill-posed or where multiple inputs can lead to the same output, as it leverages the forward model's representation to guide the inverse predictions toward feasible and meaningful solutions.

The training details and the hyperparameters of both $\mathbf{F_{DNN}}$ and the $\mathbf{I_{DNN}}$ are presented in App. \ref{app:dnnhyp}.

\section{Benchmarks and Numerical Experiments}
\label{sec:benchmarks_num_exp}

In this section, we define the inverse design benchmark problems used to analyze the performance of the active learning sampling and evaluate the accuracy of the $\mathbf{I_{DNN}}$. We provide detailed descriptions of the numerical experiments conducted, including the sampling algorithms employed for comparison with the active learning approach. Finally, we outline the accuracy metrics used to assess the performance of each method and describe the inverse design validation procedure.

\subsection{Inverse Design Benchmarks}
\label{sec:inverse_design_benchmarks}

\begin{figure}[H]
	\centering
	\begin{subfigure}[b]{1\textwidth}
		\includegraphics[width=\linewidth]{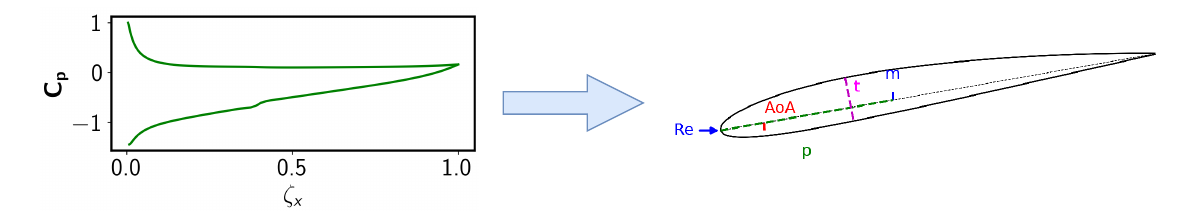}    \caption{}
		\label{fig:Figure2a}
	\end{subfigure}
	\begin{subfigure}[b]{1\textwidth}
		\includegraphics[width=\linewidth]{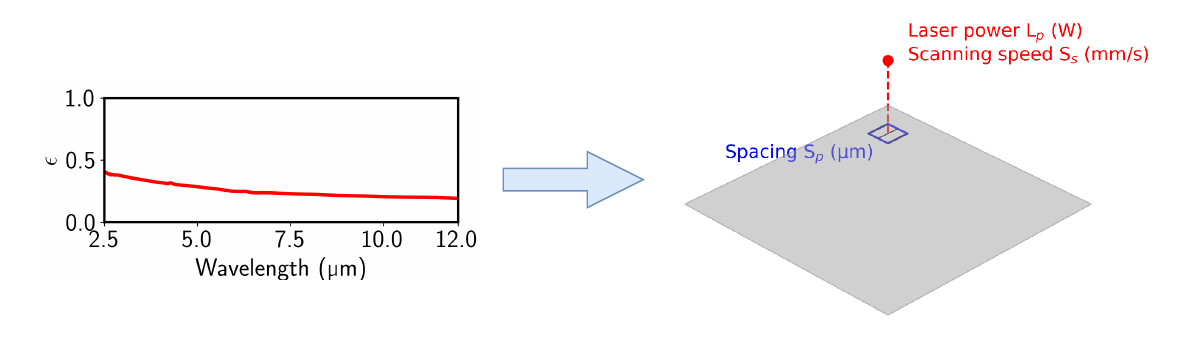}     
		\caption{}
		\label{fig:Figure2b} 
	\end{subfigure}
	\begin{subfigure}[b]{1\textwidth}
		\includegraphics[width=\linewidth]{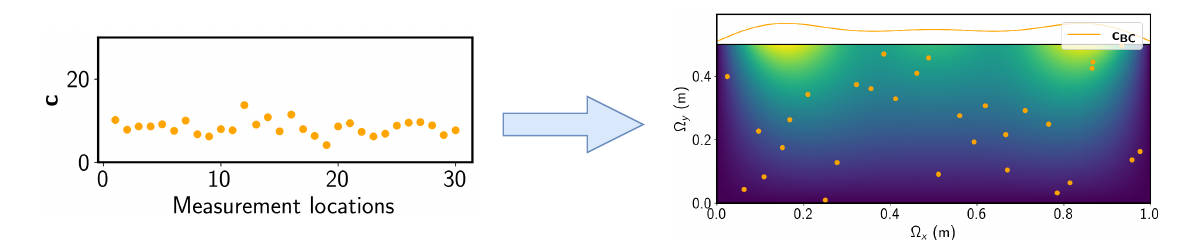}
		\caption{}
		\label{fig:Figure2c}
	\end{subfigure}
	\caption[]{Inverse design benchmark problems: (a) Airfoil inverse design (AID). (b) Photonic surfaces inverse design (PSID). (c) Scalar boundary reconstruction (SBR).}
	\label{fig:Figure2}
\end{figure}

In order to assess the performance of the active learning enhanced TNN, we develop a benchmark suite of three inverse design problems. The outline of the three problems is presented in Fig. \ref{fig:Figure2}. 

The first problem is the Airfoil Inverse Design (AID), where the goal is to determine both the flow and the design parameters of an airfoil based on its pressure coefficient distribution. In Fig. \ref{fig:Figure2a} the specifics of the AID problem are visualized. The dimensionless pressure coefficient curves are denoted as $\mathbf{C_p}$, while the flow and the geometrical parameters include the Reynolds number (Re), the Angle of Attack ($\alpha$), the maximum camber distance (m), position of the maximum camber (p), and the maximum thickness of the airfoil (t).

The second problem is the Photonic Surface Inverse Design (PSID), which aims to obtain laser manufacturing parameters corresponding to a desired spectral emissivity curve of the photonic surface (a metamaterial), visualized in Fig. \ref{fig:Figure2b}. The photonic surfaces are created by texturing the surface of the plain material (in our case the alloy Inconel) with ultra-fast lasers. The spectral emissivity curves are defined as $\mathbf{\epsilon}$, and the laser manufacturing parameters are laser power ($L_p$) (W), scanning speed ($S_s$) (mm/s), and the spacing of the textures on the surface of the metamaterial ($S_p$) ($\mu$m). 

The third problem is the Scalar Boundary Reconstruction (SBR), where the objective is to recover the boundary conditions ($\mathbf{c_{BC}}$) of a scalar diffusion PDE using scattered measurements of the scalar field ($\mathbf{c}$) within a two-dimensional domain, as shown in Fig. \ref{fig:Figure2c}. The scalar diffusion PDE models phenomena such as heat transfer or contaminant diffusion in a medium. The boundary conditions $\mathbf{c_{BC}}$ represent the values of the scalar field (e.g., temperature or concentration) along the top boundary of the domain, which are unknown and need to be determined. The scattered measurements $\mathbf{c}$ are obtained at 30 interior points within the domain and provide partial information about the scalar field's distribution. These measurements are strategically placed to capture the behavior of the field within the domain. Both $\mathbf{c_{BC}}$ and $\mathbf{c}$ are dimensionless quantities, normalized to facilitate computational analysis.

A mathematical summary of the benchmarks—including the size of the design spaces and formal definitions of the design vectors and objectives—is provided in Table \ref{tab:math_summary}. Comprehensive details of all benchmarks, such as the mathematical and numerical background, simulation software used, experimental setup and datasets, as well as the training and validation details of the HF surrogates $\mathbf{H}$, are given in Appendix \ref{app:benchmarks}.

\begin{table}[h]
	\centering
	\caption{Summary of Benchmark Problems}
	\label{tab:benchmark_summary}
	\begin{tabularx}{1.1\textwidth}{|p{3cm}|>{\centering\arraybackslash}X|>{\centering\arraybackslash}X|>{\centering\arraybackslash}X|}
		\hline
		\textbf{Aspect} & \textbf{Benchmark 1} & \textbf{Benchmark 2} & \textbf{Benchmark 3} \\
		\hline
		\textbf{Problem Name} & AID & PSID & SBR \\
		\hline
		\textbf{Design Vector \( \mathbf{x} \)} & 
		\( \mathbf{x} = \begin{bmatrix} \text{Re} \\ \alpha \\ m \\ p \\ t \end{bmatrix} \in \mathbb{R}^5 \) & 
		\( \mathbf{x} = \begin{bmatrix} \text{L}_p \\ \text{S}_s \\ \text{S}_p \end{bmatrix} \in \mathbb{R}^3 \) & 
		\( \mathbf{x} = \begin{bmatrix}
			c_{\text{BC}1} \\
			c_{\text{BC}2} \\
			\vdots \\
			c_{\text{BC}20}
		\end{bmatrix} \in \mathbb{R}^{20} \) \\
		\hline
		\textbf{Output \( \mathbf{y} \)} & 
		\( \mathbf{C_p} \in \mathbb{R}^{75} \) & 
		\( \boldsymbol{\epsilon} \in \mathbb{R}^{822} \) & 
		\( \mathbf{c} \in \mathbb{R}^{30} \) \\
		\hline
		\textbf{Forward Mapping} & 
		\( f: \mathbb{R}^5 \rightarrow \mathbb{R}^{75} \) & 
		\( f: \mathbb{R}^{3} \rightarrow \mathbb{R}^{822} \) & 
		\( f: \mathbb{R}^{20} \rightarrow \mathbb{R}^{30} \) \\
		\hline
		\textbf{Inverse Mapping} & 
		\( f^{-1}: \mathbb{R}^{75} \rightarrow \mathbb{R}^5 \) & 
		\( f^{-1}: \mathbb{R}^{822} \rightarrow \mathbb{R}^3 \) & 
		\( f^{-1}: \mathbb{R}^{30} \rightarrow \mathbb{R}^{20} \) \\
		\hline
		\textbf{Objective} & 
		Recover \( \mathbf{x} \) from \( \mathbf{C_p} \) & 
		Recover \( \mathbf{x} \) from \( \boldsymbol{\epsilon} \) & 
		Recover \( \mathbf{x} \) from \( \mathbf{c} \) \\
		\hline
	\end{tabularx}
    \label{tab:math_summary}
\end{table}

\subsection{Numerical Experiments Setup}
\label{sec:num_exp_setup}

In this section, we outline the setup of the numerical experiments conducted. Specifically, we detail the hyperparameters of the active learning process, and the sampling algorithms employed for comparison with the active learning approach.

In the active learning procedure, we employed two different algorithms to efficiently train the forward model $\mathbf{M}$: the Random Forest (RF) algorithm and an ensemble of DNNs, referred to as Deep Ensembles (DE). Both algorithms were chosen because they provide uncertainty quantification capabilities essential for active learning. Implementation details and hyperparameters for both algorithms are provided in App. \ref{app:hyperparams}. We initially used only the RF algorithm to explore their performance across different benchmark problems. However, we found that it failed to accurately model the forward relationship in the SBR benchmark problem. Consequently, we selected the DE algorithm for the SBR benchmark, as outlined in Tab. \ref{tab:num_exp}. 

Furthermore, the maximum number of samples generated by the active learning procedure varied per benchmark as outlined in Tab. \ref{tab:num_exp}. The number of maximum samples corresponds to the number of times the HF surrogate $\mathbf{H}$ is queried for a response in order to generate a dataset and train the TNN components $\mathbf{F_{DNN}}$ and $\mathbf{I_{DNN}}$. For all benchmarks, the active learning process was initialized with $n_0=$ 20 samples generated using the Latin Hypercube Sampling (LHS) algorithm, and a batch size $k=$  5 was used in each iteration of the process. To determine the optimal design vectors ($\mathbf{x}^*_i$) (defined in Alg. \ref{alg:AL}) we utilize the Particle Swarm Optimization  (PSO) algorithm (implemented in the Indago 0.5.1 Python module for optimization), where the maximum number of evaluations is set to 100, while all other hyperparameters of the PSO are set as the default recommended variables of the module (\cite{Indago}).

\begin{table}[ht]
	\centering
	\caption[]{Active learning numerical experiments setup for each benchmark.}
	\begin{tabular}{lccc}
		\hline
		\textbf{Benchmark} & \textbf{Algorithm} & \textbf{$N_{max}$} \\
		\hline
		AID & RF & 150 \\
		PSID & RF & 300 \\
		SBR & DE & 400 \\
		\hline
	\end{tabular}
	\label{tab:num_exp}
\end{table}

We employed four different sampling algorithms to generate efficient datasets for training each component of the TNN and to compare their performance with the active learning approach. The algorithms used are Random sampling (R), LHS, GreedyFP (GFP) \cite{gonzalez1985clustering, eldar1997farthest}, and Best Candidate (BC) \cite{mitchell1991spectrally} sampling. 

The BC and GFP sampling algorithms were selected as they were determined to perform efficiently for an array of tasks such as surrogate modeling, hyperparameter optimization, and data analysis \cite{kamath2022intelligent}. Both the GFP algorithm and the BC algorithm generate sequences of samples by iteratively proposing candidate samples and selecting the one that is farthest from the set of samples selected so far; however, the key difference is that GFP uses a constant number of candidate samples at each iteration, while the BC algorithm generates an increasing number of random candidate samples with each iteration. Both sampling algorithms were implemented as outlined in the work by \cite{kamath2022intelligent}. 

For each benchmark problem, the number of samples generated by these samplers is equal to the maximum number of samples specified in Tab. \ref{tab:num_exp} to ensure a fair comparison with the active learning approach. The generated samples were evaluated using the HF surrogate \( \mathbf{H} \) to form the training dataset. To account for randomness, the active learning procedure, all of the sampling algorithms used to generate the datasets, and the training of the TNN components were each repeated 30 times. The results from these repetitions were then statistically analyzed.

\subsection{Accuracy Metrics}

For the forward model ($\textbf{M}$) assessment, and the $\mathbf{I_{DNN}}$, we use the root mean square error (RMSE) (Eq. \ref{eqn:rmse}), and the coefficient of determination (R$^2$) (Eq. \ref{eqn:r2}) as they are defined in \cite{borchani2015survey}, for multioutput regression problems since all of our inverse design benchmarks fit into this category. Moreover, we also utilize the normalized maximum absolute prediction error (NMAE) \cite{surjanovic2019adaptive} (Eq. \ref{eqn:nmae}) extended for multi-output regression.

Firstly, the RMSE is defined as:

\begin{equation}
	\text{RMSE} = \sqrt{ \frac{1}{np} \sum_{i=1}^{n} \sum_{j=1}^{p} \left( y_{ij} - \hat{y}_{ij} \right)^2 }
	\label{eqn:rmse}
\end{equation}

where $n$ is the number of samples in the output test data (T$_y$), $p$ is the number of output variables (i.e. dimension of the output vectors of each inverse design benchmark problem), $y_{ij}$ is the true value of the $j$-th output for the $i$-th sample, $\hat{y}_{ij}$ is the model (either $\mathbf{M}$ or $\mathbf{I_{DNN}}$) predicted value of the $j$-th output for the $i$-th sample. Secondly, we define the R$^2$ as:

\begin{equation}
	R^2 = 1 - \frac{ \frac{1}{p} \sum\limits_{j=1}^{p} \sum\limits_{i=1}^{n} \left( y_{ij} - \hat{y}_{ij} \right)^2 }{ \frac{1}{p} \sum\limits_{j=1}^{p} \sum\limits_{i=1}^{n} \left( y_{ij} - \bar{y}_j \right)^2 }
	\label{eqn:r2}
\end{equation}

where $\bar{y}_j$ is the mean of the true values for output $j$ defined as $\bar{y}_j = \frac{1}{n} \sum_{i=1}^{n} y_{ij}$. Finally, we define the NMAE as:

\begin{equation}
	\text{NMAE} = \frac{1}{p} \sum_{j=1}^{p} \frac{ \displaystyle \max_{1 \leq i \leq n} \left| y_{ij} - \hat{y}_{ij} \right| }{ \displaystyle \max_{1 \leq i \leq n} \left| y_{ij} - \bar{y}_j \right| }
	\label{eqn:nmae}
\end{equation}

By employing these metrics, we can comprehensively evaluate the performance of our models across all output variables, ensuring that both average performance (through RMSE and R$^2$) and worst-case scenarios (through NMAE) are adequately assessed.

\subsection{Inverse Design Validation}

\begin{figure}[h]
	\centering
	\includegraphics[width=\textwidth]{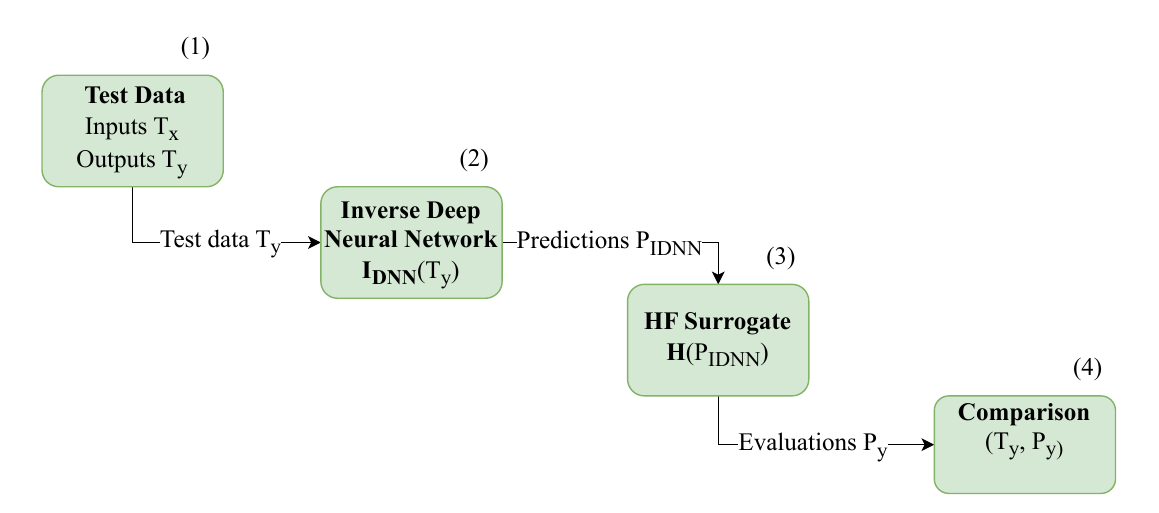}
	\caption{Inverse design validation procedure aiming to assess the accuracy of the trained \( \mathbf{I_{DNN}} \) using the test data from each inverse design benchmark problem. In step (1), we define the test dataset comprising input-output pairs (T$_x$, T$_y$). In step (2), we utilize the output values T$_y$ as inputs to the inverse model \( \mathbf{I_{DNN}} \) to obtain the predicted inputs P$_{\text{IDNN}}$. In step (3), we reconstruct the outputs P$_y$ by feeding the predicted inputs P$_{\text{IDNN}}$ into the HF surrogate \( \mathbf{H} \). Finally, in step (4), we compare the original output values T$_y$ from the test dataset with the reconstructed outputs P$_y$ to evaluate the inverse model's accuracy using the specified metrics.}
	\label{fig:Figure3}
\end{figure}

To assess the accuracy metrics (RMSE, R$^2$, and NMAE) of the $\mathbf{I_{DNN}}$ on the test data, we utilize the validation framework presented in Fig. \ref{fig:Figure3}. Specifically, since we are interested in the inverse relationship, we use the output values T$_y$ from the test dataset as inputs to the $\mathbf{I_{DNN}}$. The $\mathbf{I_{DNN}}$ then produces predictions of the original inputs, which we denote as P$_{IDNN}$.

Next, we feed these predicted inputs P$_{IDNN}$ into the HF surrogate $\mathbf{H}$ to generate reconstructed outputs, denoted as P$_y$. Essentially, P$_y$ are the outputs that the model $\mathbf{H}$ would produce given the predicted inputs from the $\mathbf{I_{DNN}}$. We then compare these reconstructed outputs P$_y$ with the actual output values T$_y$ from the test dataset using the accuracy metrics. This comparison allows us to evaluate how well the $\mathbf{I_{DNN}}$, in conjunction with the HF surrogate, can reproduce the original outputs, thereby assessing the $\mathbf{I_{DNN}}$ model's performance. 

In all inverse design benchmarks, we utilized a test dataset (T$_x$, T$_y$) consisting of 1,000 randomly selected instances (also unseen by the HF surrogate $\mathbf{H}$). The $\mathbf{I_{DNN}}$ was trained using the maximum number of samples specified in Tab. \ref{tab:num_exp}, which was significantly fewer than 1,000 in each case. Consequently, the size of the training set was much smaller than that of the test set, resulting in a train/test size ratio heavily skewed towards the test set. This imbalance highlights the challenge for $\mathbf{I_{DNN}}$ to generalize effectively from a limited amount of training data to a larger, diverse test set. Further details of the test dataset for each inverse design benchmark problem are available in App. \ref{app:benchmarks}.

\newpage
\section{AutoTandemML Framework Results}\label{sec:results}

In this section, we present the results of the AutoTandemML framework applied to all three inverse design benchmark problems. To assess the performance of the active learning approach, we also compare these results with those obtained when the dataset used to train the TNN is generated using other sampling methods (R, LHS, GFP, and BC). 

\subsection{Airfoil Inverse Design Results}

\begin{figure}[H]
\centering
\begin{subfigure}[b]{0.49\textwidth}
   \includegraphics[width=\linewidth]{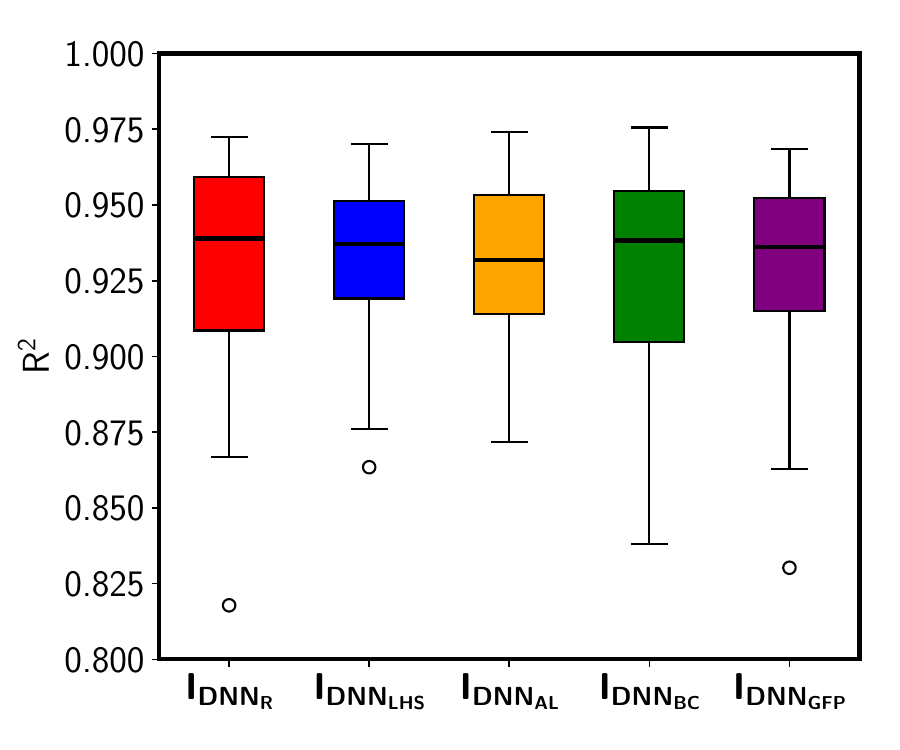}    \caption{}
   \label{fig:Figure4a}
\end{subfigure}
\begin{subfigure}[b]{0.49\textwidth}
   \includegraphics[width=\linewidth]{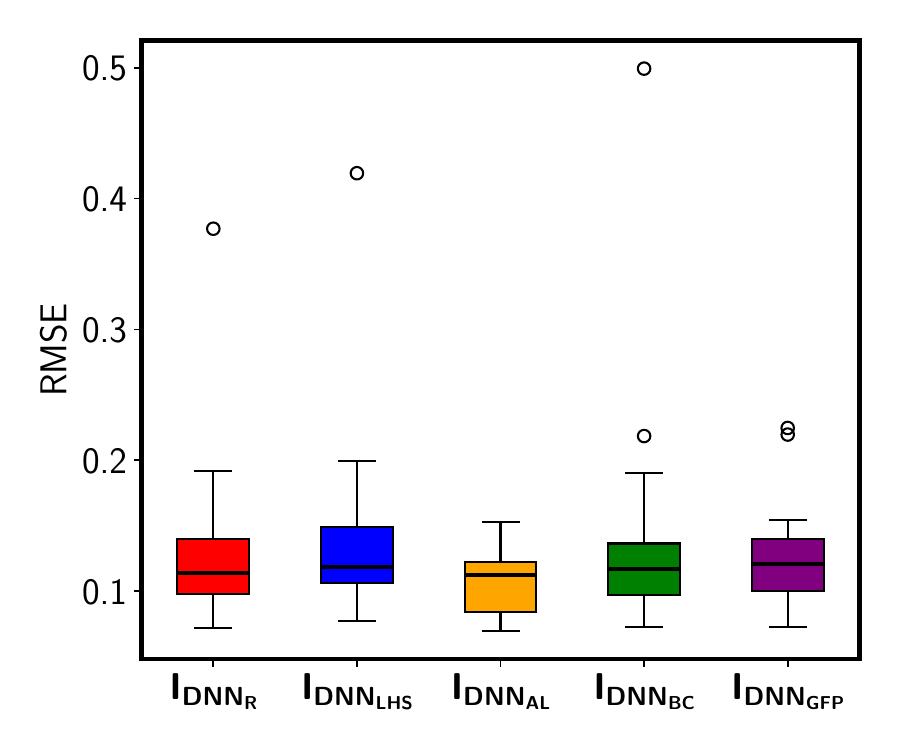}     
   \caption{}
   \label{fig:Figure4b} 
\end{subfigure}
\begin{subfigure}[b]{0.49\textwidth}
   \includegraphics[width=\linewidth]{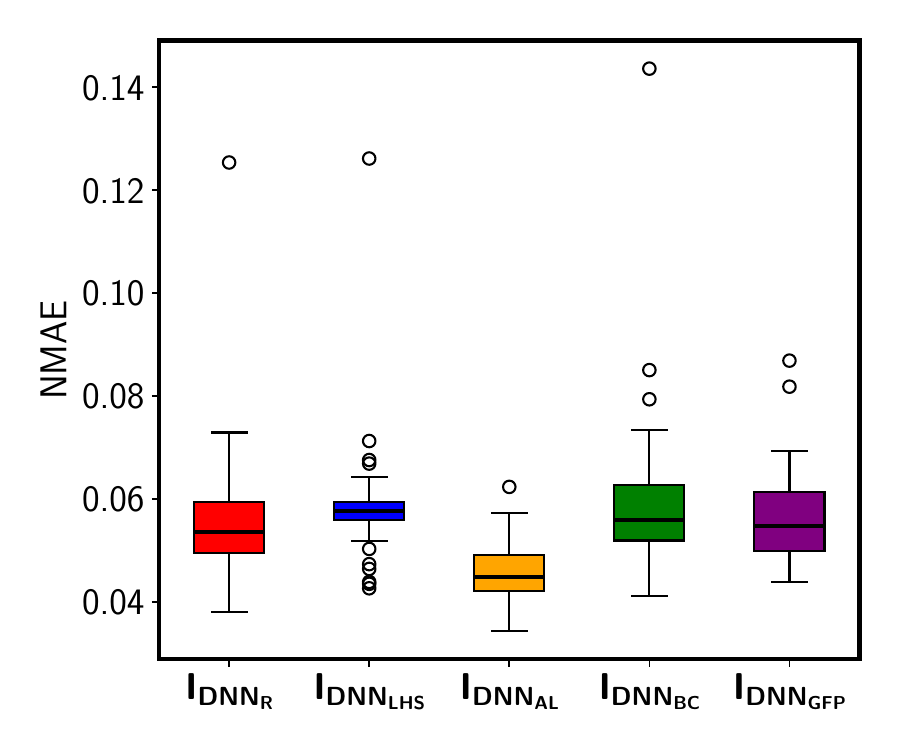}
   \caption{}
   \label{fig:Figure4c}
\end{subfigure}
\caption[Performance metrics for inverse DNN with different sampling methods]{Inverse DNN ($\mathbf{I_{DNN}}$) performance on AID benchmark problem using different dataset generation methods: (a) R$^2$ (higher is better), (b) RMSE (lower is better), and (c) NMAE (lower is better). Subscripts in $\mathbf{I_{DNN}}$ denote the sampling method (e.g., $\mathbf{I_{DNN_R}}$ for random sampling).}
\label{fig:Figure4}
\end{figure}

Fig. \ref{fig:Figure4} presents the results of the AID benchmark. Specifically, the R$^2$ metric results are shown in Fig. \ref{fig:Figure4a}, the RMSE results are displayed in Fig. \ref{fig:Figure4b}, and the NMAE results are illustrated in Fig. \ref{fig:Figure4c}. All three metrics are represented as box plots as they provide a concise visual summary of each metric's distribution by displaying their medians, quartiles, and outliers. The results are also summarized statistically in Tab. \ref{tab:aid_summary}.

Fig. \ref{fig:Figure4a} indicates that all approaches perform similarly in terms of the R$^2$ score. However, Tab. \ref{tab:aid_summary} reveals that the active learning approach ($\mathbf{I_{DNN_{AL}}}$) outperforms other inverse DNNs trained with different samplers when considering the mean R$^2$ score (R$^2$=0.93). Furthermore, the active learning approach exhibits the best performance regarding outliers of the obtained R$^2$ scores, as the minimum R$^2$ from the 30 runs is R$^2$=0.87, the close second best approach in terms of the R$^2$ score is $\mathbf{I_{DNN_{GFP}}}$. Some models obtained a negative R$^2$ in as their worst performing run, indicating that their predictions did not explain any of the variability in the data and are inferior to just using the mean of the target variable as a predictor.

In Fig. \ref{fig:Figure4b}, we observe that when considering RMSE, the $\mathbf{I_{DNN_{AL}}}$ also performs the best among the other samplers. All other approaches exhibit outliers that increase the overall RMSE. In Tab. \ref{tab:aid_summary}, while the mean RMSE of the $\mathbf{I_{DNN_{AL}}}$ is the best, the significant difference between the active learning approach and other sampling approaches becomes apparent when examining the maximum RMSE values. Specifically, the maximum RMSE for $\mathbf{I_{DNN_{AL}}}$ is 0.1529, whereas, as a reference, the worst maximum RMSE is 0.4993 for $\mathbf{I_{DNN_{BC}}}$.

Finally, in Fig.\ \ref{fig:Figure4c}, the NMAE metric confirms that $\mathbf{I_{DNN_{AL}}}$ also outperforms the other approaches, achieving the lowest maximum NMAE (0.0459). This significant difference in maximum NMAE is further evident when comparing $\mathbf{I_{DNN_{AL}}}$ to other approaches, as observed in Tab. \ref{tab:aid_summary}, i.e. the maximum NMAE for $\mathbf{I_{DNN_{AL}}}$ is 0.0623, which is close to the mean NMAE of the $\mathbf{I_{DNN_{BC}}}$ (0.0604). Moreover, not only does $\mathbf{I_{DNN_{AL}}}$ perform best when considering the mean of all metrics, but it also exhibits the least uncertainty across the 30 runs, achieving the lowest standard deviation for all metrics (third column of Tab. \ref{tab:aid_summary}). The details of the forward model ($\mathbf{M}$) performance on the AID benchmark can be found in App. \ref{app:forward_model_aid}.

\begin{table}[ht]
	\centering
	\caption[Statistical analysis of inverse DNN performance]{Statistical analysis of the $\mathbf{I_{DNN}}$ performance on AID benchmark problem using different dataset generation methods. Bold values indicate best performance in each metric. RMSE and NMAE values should be as low as possible, while R$^2$ should be as high as possible.}
	\begin{tabular}{lccccc}
		\hline
		\textbf{Metric} & \textbf{Method} & \textbf{Mean} & \textbf{Std} & \textbf{Max} & \textbf{Min} \\
		\hline
RMSE & $\mathbf{I_{DNN_R}}$ & 0.1261 & 0.0539 & 0.3770 & 0.0717 \\
RMSE & $\mathbf{I_{DNN_{LHS}}}$ & 0.1341 & 0.0605 & 0.4194 & 0.0770 \\
RMSE & $\mathbf{I_{DNN_{AL}}}$ & \textbf{0.1086} & \textbf{0.0239} & \textbf{0.1529} & \textbf{0.0693} \\
RMSE & $\mathbf{I_{DNN_{BC}}}$ & 0.1313 & 0.0752 & 0.4993 & 0.0724 \\
RMSE & $\mathbf{I_{DNN_{GFP}}}$ & 0.1244 & 0.0341 & 0.2247 & 0.0724 \\
\hline
R$^2$ & $\mathbf{I_{DNN_R}}$ & 0.9174 & 0.0915 & 0.9725 & 0.4592 \\
R$^2$ & $\mathbf{I_{DNN_{LHS}}}$ & 0.8878 & 0.2439 & 0.9702 & -0.4185 \\
R$^2$ & $\mathbf{I_{DNN_{AL}}}$ & \textbf{0.9319} & \textbf{0.0254} & 0.9742 & \textbf{0.8719} \\
R$^2$ & $\mathbf{I_{DNN_{BC}}}$ & 0.8526 & 0.4169 & \textbf{0.9756} & -1.3843 \\
R$^2$ & $\mathbf{I_{DNN_{GFP}}}$ & 0.9274 & 0.0386 & 0.9683 & 0.7920 \\
\hline
NMAE & $\mathbf{I_{DNN_R}}$ & 0.0569 & 0.0150 & 0.1254 & 0.0380 \\
NMAE & $\mathbf{I_{DNN_{LHS}}}$ & 0.0589 & 0.0142 & 0.1261 & 0.0426 \\
NMAE & $\mathbf{I_{DNN_{AL}}}$ & \textbf{0.0459} & \textbf{0.0067} & \textbf{0.0623} & \textbf{0.0343} \\
NMAE & $\mathbf{I_{DNN_{BC}}}$ & 0.0604 & 0.0180 & 0.1436 & 0.0411 \\
NMAE & $\mathbf{I_{DNN_{GFP}}}$ & 0.0574 & 0.0102 & 0.0869 & 0.0439 \\
		\hline
	\end{tabular}
\label{tab:aid_summary}
\end{table}

\newpage
\subsection{Photonic Surface Inverse Design Results}

\begin{figure}[H]
\centering
\begin{subfigure}[b]{0.49\textwidth}
   \includegraphics[width=\linewidth]{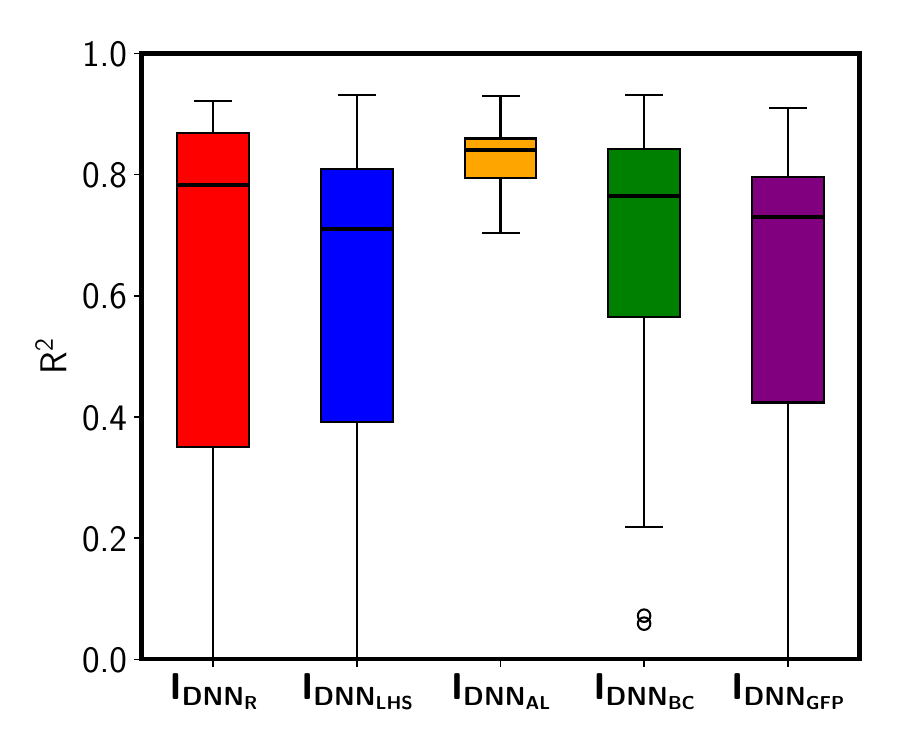}    \caption{}
   \label{fig:Figure5a}
\end{subfigure}
\begin{subfigure}[b]{0.49\textwidth}
   \includegraphics[width=\linewidth]{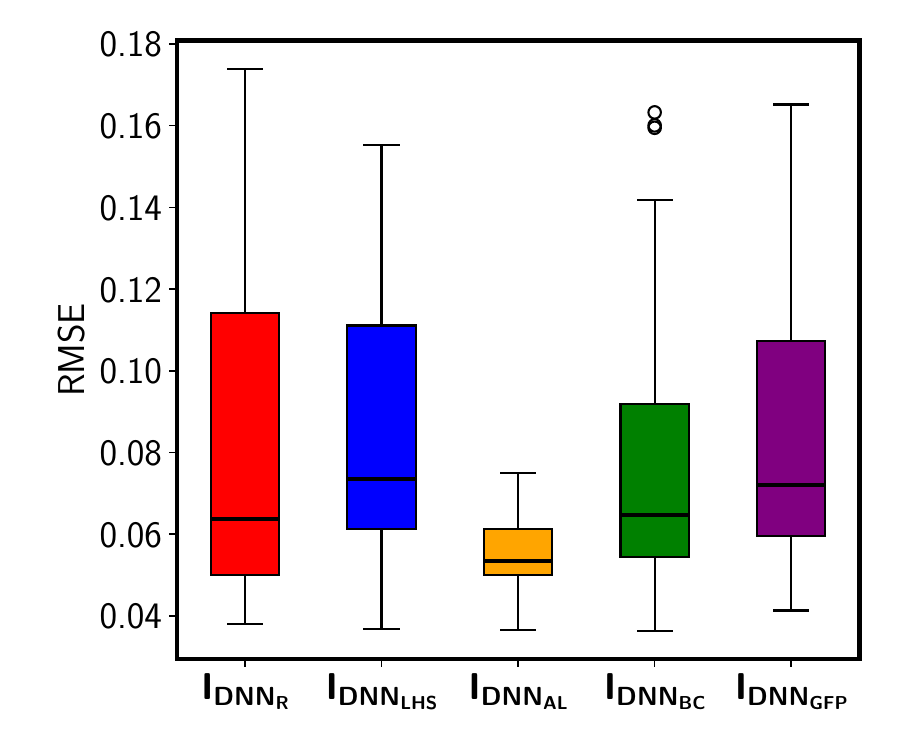}     
   \caption{}
   \label{fig:Figure5b} 
\end{subfigure}
\begin{subfigure}[b]{0.49\textwidth}
   \includegraphics[width=\linewidth]{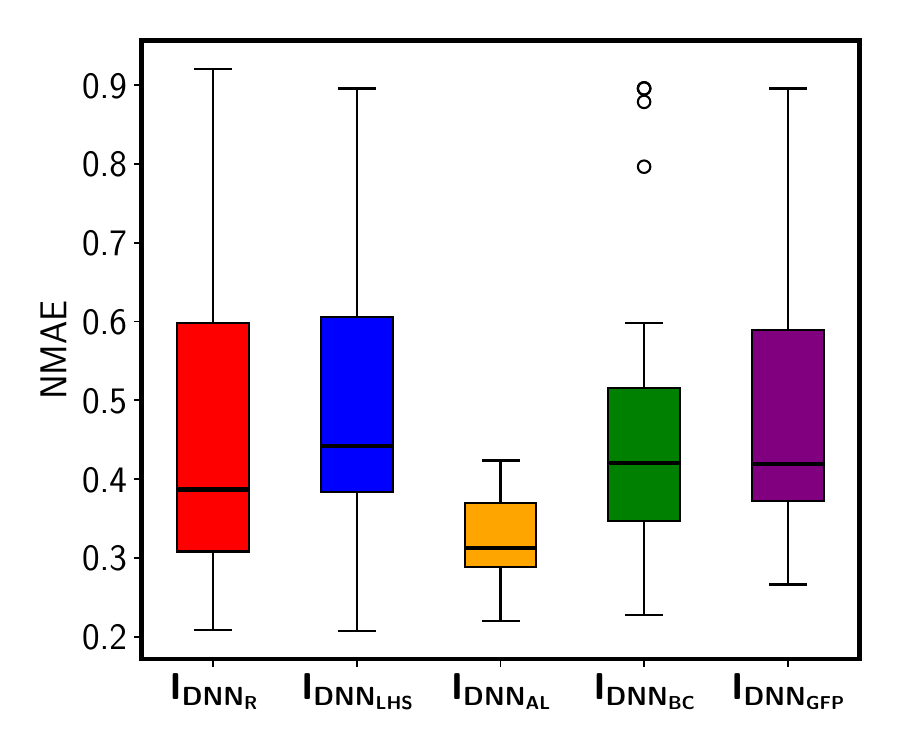}
   \caption{}
   \label{fig:Figure5c}
\end{subfigure}
\caption[Performance metrics for inverse DNN with different sampling methods]{$\mathbf{I_{DNN}}$ performance on PSID benchmark problem using different dataset generation methods: (a) R$^2$ (higher is better), (b) RMSE (lower is better), and (c) NMAE (lower is better). Subscripts in $\mathbf{I_{DNN}}$ denote the sampling method (e.g., $\mathbf{I_{DNN_R}}$ for random sampling).}
\label{fig:Figure5}
\end{figure}

Fig. \ref{fig:Figure5} presents the results of the PSID benchmark. The R$^2$ metric is shown in Fig.\ \ref{fig:Figure5a}, the RMSE results are displayed in Fig.\ \ref{fig:Figure5b}, and the NMAE results are illustrated in Fig.\ \ref{fig:Figure5c}. The results are also statistically summarized in Tab.\ \ref{tab:psid_summary}.

Fig. \ref{fig:Figure5a} indicates that $\mathbf{I_{DNN_{AL}}}$ outperforms all other sampling methods in terms of the R$^2$ score, and that it exhibits a very narrow interquartile range when compared to other samplers. Tab.\ \ref{tab:psid_summary} reveals that the difference between $\mathbf{I_{DNN_{AL}}}$ and the inverse DNNs trained with other samplers is substantial when considering the mean R$^2$ scores. Specifically, the mean R$^2$ for $\mathbf{I_{DNN_{AL}}}$ is 0.82, while the second-best performing approach, $\mathbf{I_{DNN_{BC}}}$, has an R$^2$ of 0.60.

In Fig.\ \ref{fig:Figure5b}, we observe that when considering RMSE, $\mathbf{I_{DNN_{AL}}}$ also performs the best among the sampling methods and exhibits the narrowest interquartile range. All other approaches display increased RMSE values. In Tab.\ \ref{tab:psid_summary}, it can be seen that both the mean RMSE and the maximum RMSE of $\mathbf{I_{DNN_{AL}}}$ are the lowest compared to the other sampling approaches. Specifically, the mean and maximum RMSE for $\mathbf{I_{DNN_{AL}}}$ are 0.056 and 0.075, respectively. The second-best performing approach, $\mathbf{I_{DNN_{BC}}}$, obtains mean and maximum RMSE values of 0.080 and 0.16, respectively.

Finally, in Fig.\ \ref{fig:Figure5c}, the NMAE metric confirms that $\mathbf{I_{DNN_{AL}}}$ outperforms the other approaches, achieving the lowest mean NMAE of 0.324. This significant difference in NMAE is further evident when comparing $\mathbf{I_{DNN_{AL}}}$ to other approaches, as shown in Tab. \ref{tab:psid_summary}; specifically, the maximum NMAE for $\mathbf{I_{DNN_{AL}}}$ is 0.42, while the worst-performing approach, $\mathbf{I_{DNN_{R}}}$, has a maximum NMAE of 0.92. Additionally, $\mathbf{I_{DNN_{AL}}}$ exhibits the narrowest interquartile range in the NMAE metric, reflecting its consistent performance across the 30 runs.

Moreover, $\mathbf{I_{DNN_{AL}}}$ performs the best when considering the mean of all metrics and exhibits the least uncertainty across the 30 runs, achieving the lowest standard deviation for all metrics (third column of Tab.\ \ref{tab:psid_summary}). The details of the forward model ($\mathbf{M}$) performance on the PSID benchmark can be found in App. \ref{app:forward_model_psid}.

\begin{table}[ht]
	\centering
	\caption[Statistical analysis of inverse DNN performance]{Statistical analysis of the $\mathbf{I_{DNN}}$ performance on PSID benchmark problem using different dataset generation methods. Bold values indicate best performance in each metric. RMSE and NMAE values should be as low as possible, while R$^2$ should be as high as possible.}
	\begin{tabular}{lccccc}
		\hline
		\textbf{Metric} & \textbf{Method} & \textbf{Mean} & \textbf{Std} & \textbf{Max} & \textbf{Min} \\
		\hline
RMSE & $\mathbf{I_{DNN_R}}$ & 0.0820 & 0.0426 & 0.1740 & 0.0381 \\
RMSE & $\mathbf{I_{DNN_{LHS}}}$ & 0.0838 & 0.0339 & 0.1552 & 0.0369 \\
RMSE & $\mathbf{I_{DNN_{AL}}}$ & \textbf{0.0560} & \textbf{0.0092} & \textbf{0.0750} & 0.0365 \\
RMSE & $\mathbf{I_{DNN_{BC}}}$ & 0.0806 & 0.0380 & 0.1632 & \textbf{0.0362} \\
RMSE & $\mathbf{I_{DNN_{GFP}}}$ & 0.0873 & 0.0386 & 0.1652 & 0.0413 \\
\hline
R$^2$ & $\mathbf{I_{DNN_R}}$ & 0.5747 & 0.4141 & 0.9214 & -0.3864 \\
R$^2$ & $\mathbf{I_{DNN_{LHS}}}$ & 0.5898 & 0.3144 & 0.9309 & -0.1665 \\
R$^2$ & $\mathbf{I_{DNN_{AL}}}$ & \textbf{0.8224} & \textbf{0.0562} & 0.9299 & \textbf{0.7042} \\
R$^2$ & $\mathbf{I_{DNN_{BC}}}$ & 0.6094 & 0.3455 & \textbf{0.9315} & -0.1932 \\
R$^2$ & $\mathbf{I_{DNN_{GFP}}}$ & 0.5589 & 0.3519 & 0.9093 & -0.2343 \\
\hline
NMAE & $\mathbf{I_{DNN_R}}$ & 0.4701 & 0.2136 & 0.9209 & 0.2085 \\
NMAE & $\mathbf{I_{DNN_{LHS}}}$ & 0.5000 & 0.1902 & 0.8957 & \textbf{0.2069} \\
NMAE & $\mathbf{I_{DNN_{AL}}}$ & \textbf{0.3244} & \textbf{0.0500} & \textbf{0.4236} & 0.2197 \\
NMAE & $\mathbf{I_{DNN_{BC}}}$ & 0.4823 & 0.1945 & 0.8957 & 0.2280 \\
NMAE & $\mathbf{I_{DNN_{GFP}}}$ & 0.5136 & 0.2117 & 0.8957 & 0.2663 \\
		\hline
	\end{tabular}
\label{tab:psid_summary}
\end{table}

\newpage
\subsection{Scalar Boundary Reconstruction Results}

\begin{figure}[H]
\centering
\begin{subfigure}[b]{0.49\textwidth}
   \includegraphics[width=\linewidth]{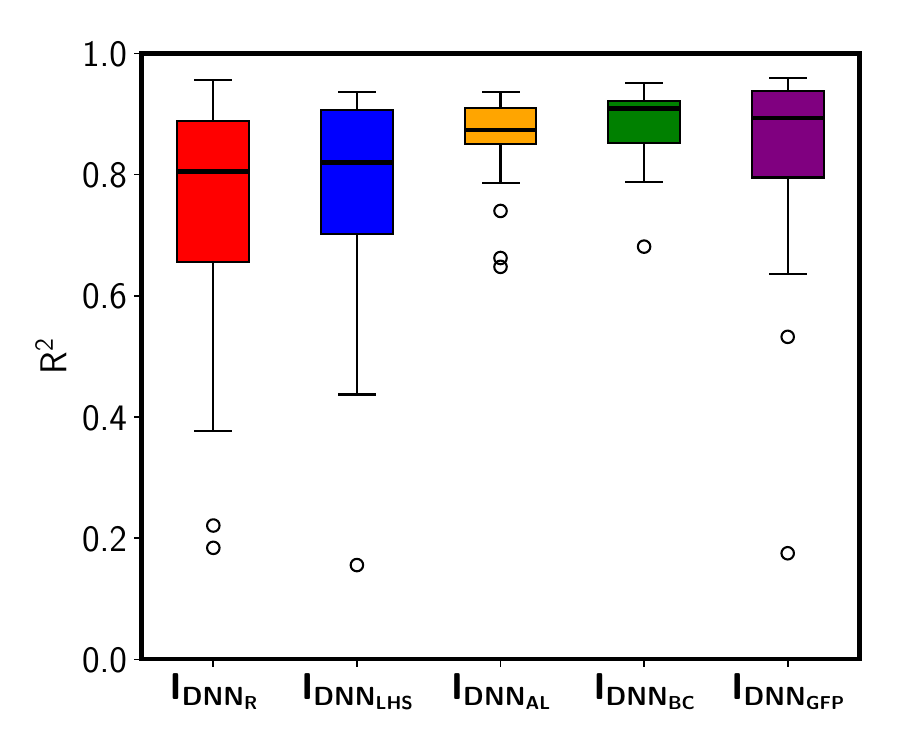}    \caption{}
   \label{fig:Figure6a}
\end{subfigure}
\begin{subfigure}[b]{0.49\textwidth}
   \includegraphics[width=\linewidth]{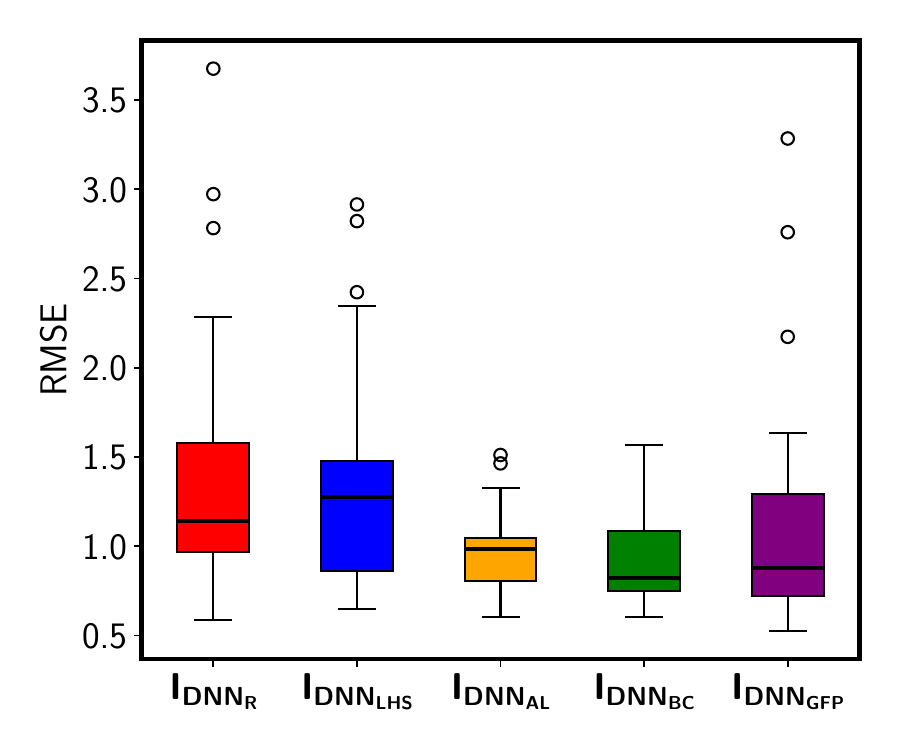}     
   \caption{}
   \label{fig:Figure6b} 
\end{subfigure}
\begin{subfigure}[b]{0.49\textwidth}
   \includegraphics[width=\linewidth]{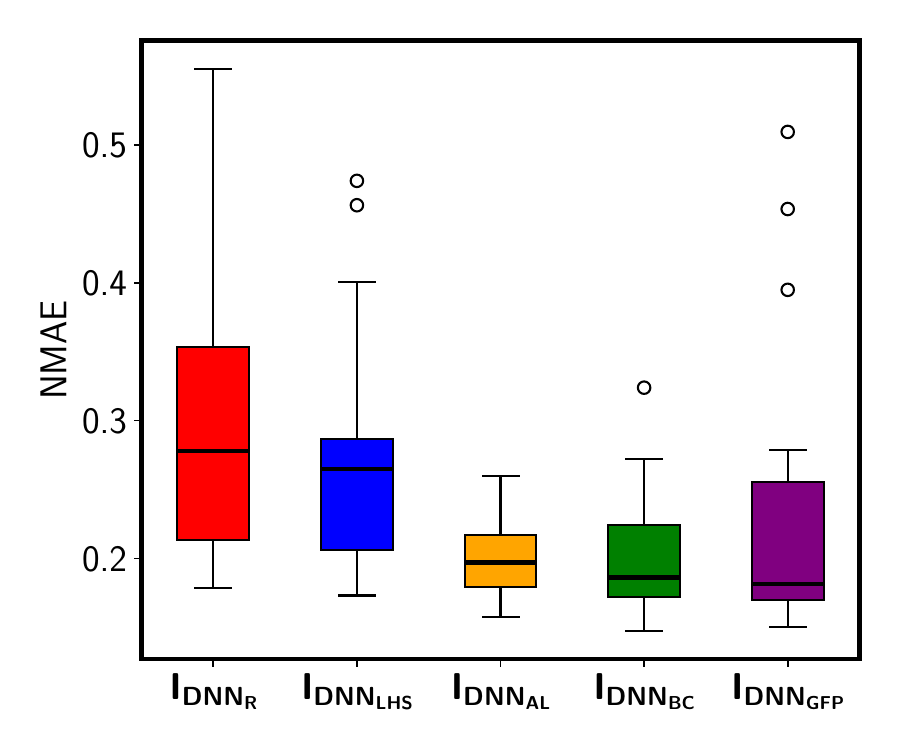}
   \caption{}
   \label{fig:Figure6c}
\end{subfigure}
\caption[Performance metrics for inverse DNN with different sampling methods]{$\mathbf{I_{DNN}}$ performance on SBR benchmark problem using different dataset generation methods: (a) R$^2$ (higher is better), (b) RMSE (lower is better), and (c) NMAE (lower is better). Subscripts in $\mathbf{I_{DNN}}$ denote the sampling method (e.g., $\mathbf{I_{DNN_R}}$ for random sampling).}
\label{fig:Figure6}
\end{figure}

Fig.\ \ref{fig:Figure6} presents the results of the SBR benchmark. The R$^2$ metric is shown in Fig.\ \ref{fig:Figure6a}, the RMSE results are displayed in Fig.\ \ref{fig:Figure6b}, and the NMAE results are illustrated in Fig.\ \ref{fig:Figure6c}. The results are also statistically summarized in Tab.\ \ref{tab:sbr_summary}.

Fig.\ \ref{fig:Figure6a} indicates that $\mathbf{I_{DNN_{BC}}}$ outperforms all other sampling methods in terms of the R$^2$ score, achieving the highest mean R$^2$ of 0.885. The second-best performing approach, $\mathbf{I_{DNN_{AL}}}$, has a mean R$^2$ of 0.860, as shown in Tab.\ \ref{tab:sbr_summary}. Although $\mathbf{I_{DNN_{AL}}}$ has a slightly lower mean R$^2$, it exhibits the narrowest interquartile range among all methods, suggesting more consistent performance across the 30 runs. 

In Fig.\ \ref{fig:Figure6b}, we observe that when considering RMSE, $\mathbf{I_{DNN_{BC}}}$ again performs the best among the sampling methods, obtaining the lowest mean RMSE of 0.905. The second-best performing approach, $\mathbf{I_{DNN_{AL}}}$, achieves a mean RMSE of 0.978. While $\mathbf{I_{DNN_{AL}}}$ has a slightly higher mean RMSE than $\mathbf{I_{DNN_{BC}}}$, it exhibits the narrowest interquartile range, indicating more consistent error rates across different runs. Finally, in Fig.\ \ref{fig:Figure6c}, the NMAE metric reveals that $\mathbf{I_{DNN_{AL}}}$ slightly outperforms the other approaches, achieving the lowest mean NMAE of 0.2002 compared to 0.2009 for $\mathbf{I_{DNN_{BC}}}$. This suggests that, in terms of normalized absolute errors, $\mathbf{I_{DNN_{AL}}}$ provides marginally better performance.

Moreover, $\mathbf{I_{DNN_{AL}}}$ demonstrates comparable performance to $\mathbf{I_{DNN_{BC}}}$ when considering the mean of all metrics and exhibits the least uncertainty across the 30 runs, achieving the lowest standard deviations for RMSE and NMAE (as shown in the third column of Tab.\ \ref{tab:sbr_summary}). Additionally, $\mathbf{I_{DNN_{AL}}}$ exhibits the narrowest interquartile range for all metrics, reflecting its consistent performance across the 30 runs. These observations highlight the effectiveness of the Active Learning sampling method in providing reliable and consistent results. The details of the forward model ($\mathbf{M}$) performance on the SBR benchmark can be found in App.\ \ref{app:forward_model_sbr}.

\begin{table}[ht]
	\centering
	\caption[Statistical analysis of inverse DNN performance]{Statistical analysis of the Inverse DNN ($\mathbf{I_{DNN}}$) performance on SBR benchmark problem using different dataset generation methods. Bold values indicate best performance in each metric. RMSE and NMAE values should be as low as possible, while R$^2$ should be as high as possible.}
	\begin{tabular}{lccccc}
		\hline
		\textbf{Metric} & \textbf{Method} & \textbf{Mean} & \textbf{Std} & \textbf{Max} & \textbf{Min} \\
		\hline
RMSE & $\mathbf{I_{DNN_R}}$ & 1.4125 & 0.7054 & 3.6755 & 0.5859 \\
RMSE & $\mathbf{I_{DNN_{LHS}}}$ & 1.3290 & 0.5998 & 2.9147 & 0.6482 \\
RMSE & $\mathbf{I_{DNN_{AL}}}$ & 0.9781 & 0.2176 & \textbf{1.5112} & 0.6016 \\
RMSE & $\mathbf{I_{DNN_{BC}}}$ & \textbf{0.9054} & \textbf{0.2133} & 1.5668 & 0.6047 \\
RMSE & $\mathbf{I_{DNN_{GFP}}}$ & 1.1048 & 0.6339 & 3.2841 & \textbf{0.5237} \\
\hline
R$^2$ & $\mathbf{I_{DNN_R}}$ & 0.7122 & 0.2880 & 0.9565 & -0.4353 \\
R$^2$ & $\mathbf{I_{DNN_{LHS}}}$ & 0.7309 & 0.2726 & 0.9366 & -0.1042 \\
R$^2$ & $\mathbf{I_{DNN_{AL}}}$ & 0.8601 & 0.0694 & 0.9367 & 0.6479 \\
R$^2$ & $\mathbf{I_{DNN_{BC}}}$ & \textbf{0.8852} & \textbf{0.0576} & 0.9513 & \textbf{0.6812} \\
R$^2$ & $\mathbf{I_{DNN_{GFP}}}$ & 0.7885 & 0.3206 & \textbf{0.9602} & -0.7087 \\
\hline
NMAE & $\mathbf{I_{DNN_R}}$ & 0.2951 & 0.0946 & 0.5554 & 0.1786 \\
NMAE & $\mathbf{I_{DNN_{LHS}}}$ & 0.2726 & 0.0791 & 0.4739 & 0.1731 \\
NMAE & $\mathbf{I_{DNN_{AL}}}$ & \textbf{0.2002} & \textbf{0.0275} & \textbf{0.2597} & 0.1575 \\
NMAE & $\mathbf{I_{DNN_{BC}}}$ & 0.2009 & 0.0411 & 0.3239 & \textbf{0.1472} \\
NMAE & $\mathbf{I_{DNN_{GFP}}}$ & 0.2230 & 0.0868 & 0.5094 & 0.1505 \\
		\hline
	\end{tabular}
\label{tab:sbr_summary}
\end{table}

\newpage
\section{Conclusion}\label{sec:conclusion}

We introduced and investigated the AutoTandemML framework for inverse design problems in science and engineering. AutoTandemML synergistically combines active learning with TNNs to efficiently generate datasets for accurate inverse design solutions. We evaluated the framework on three benchmarks—the airfoil inverse design, photonic surfaces inverse design, and scalar boundary reconstruction—and demonstrated excellent performance across all. Compared to other sampling algorithms, the TNN trained with active learning outperformed others in two benchmarks and was competitive in the third. Notably, AutoTandemML offers reliable performance with low variability across repeated experiments, a significant advantage for inverse design applications.

Future research could explore applying AutoTandemML to other inverse design problems, enhancing the active learning component with more sophisticated uncertainty quantification methods, and developing hybrid approaches that combine active learning with best candidate sampling. Additionally, extending the TNN architecture to other deep neural networks like Graph Neural Networks could enable the framework to handle discrete and graph-structured datasets, opening new possibilities in areas like molecular inverse design. Ultimately, a comprehensive scalability study should be conducted to deepen our understanding of the problem dimensionality for which this approach is most effective.

\section*{Acknowledgments}
This work was supported by the Laboratory Directed Research and Development Program of Lawrence Berkeley National Laboratory under U.S. Department of Energy Contract No. DE-AC02-05CH11231.  M\"uller's time was supported under U.S. Department of Energy Contract  No. DE-AC36-08GO28308, U.S. Department of Energy Office of Science, Office of Advanced Scientific Computing Research, Scientific Discovery
through Advanced Computing (SciDAC) program through the FASTMath Institute to the National Renewable Energy Laboratory.

\section*{Author Contributions}
L.G. wrote the manuscript, developed the methods, developed the code, designed the numerical experiments, and analyzed the performance of the algorithms, J.M. and W.A.J. supervised the research and edited the manuscript.

\section*{Declaration of Competing Interest}
The authors declare that they have no known competing financial interests or personal relationships that could have appeared to influence the work reported in this paper.

\section*{Data Availability}
The AutoTandemML code needed to reproduce the study can be found on the following repository: \url{https://github.com/lukagrbcic/AutoTandemML}. The inverse design benchmark HF surrogates, and the train/test data can be found on the following repository: \url{https://github.com/lukagrbcic/InverseBench.}
\newpage
\begin{appendices}

\section{Machine Learning Algorithms Hyperparameters and Training}\label{app:hyperparams}

In this section we present the hyperparameter and training details of the forward and inverse DNN models of the TNN. Moreover, we present all of the training details and the hyperparameters of the algorithms used for training the active learning model $\mathbf{M}$ and uncertainty quantification. 

\subsection{Forward DNN ($\mathbf{F_{DNN}}$) and Inverse DNN ($\mathbf{I_{DNN}})$}\label{app:dnnhyp}

Both the forward DNN ($\mathbf{F_{DNN}}$) and the inverse DNN ($\mathbf{I_{DNN}}$) were configured with identical hyperparameters and training settings, except for the loss functions used, and the input and output layers reversed. Each network was implemented as a multi-layer perceptron (MLP) comprising five hidden layers with neuron counts of 64, 128, 256, 128, and 64 neurons, respectively. The Rectified Linear Unit (ReLU) activation function was employed in all hidden layers to introduce non-linearity into the models. Training was conducted over 2,000 epochs with a batch size of 32 and a learning rate of 0.001, ensuring efficient and stable convergence during optimization.

Both networks utilized the Adam optimizer for weight updates, a validation split of 10\% to monitor performance on unseen data, and an early stopping mechanism with a patience parameter of 10 epochs to prevent overfitting. The RMSE loss function for both networks. All training procedures for the forward and inverse neural networks were carried out using PyTorch 2.3.0 with Python 3.10 \cite{paszke2019pytorch}.

Furthermore, prior to training, we applied Min-Max scaling to both the input features and output targets using the $\texttt{MinMaxScaler}$ function from scikit-learn version 1.2.2 \cite{pedregosa2011scikit}. This preprocessing step normalized both the inputs and outputs to a range between 0 and 1, which helped improve the training process by ensuring that all variables contribute equally to the learning process.

For the inverse DNN training, we did not rescale the data anew. Instead, we reused the Min-Max scalers from the forward DNN training, but applied them in reverse order. Specifically, the scaler that was used for the outputs in the forward DNN was applied to the inputs in the inverse DNN, and vice versa for the scaler used on the inputs. This approach maintains consistency between the forward and inverse models and ensures that the scaling corresponds appropriately to the data being modeled.

\subsection{Random Forests}\label{app:rfhyp}

RFs (\cite{breiman2001random}) were employed as the model $\mathbf{M}$ during the active learning procedure for the AID and PSID benchmark problems. RFs function by constructing an ensemble of decision trees, each trained on a bootstrap sample of the original dataset. At each node in a tree, a random subset of input features is selected to determine the best split, introducing additional randomness that helps to reduce overfitting and enhance model robustness. Each individual decision tree generates its own prediction, and these predictions are then aggregated—by averaging in regression tasks—to produce the final output. To perform uncertainty quantification with the RF model, we calculate the standard deviation of the predictions from all individual trees in the ensemble. This standard deviation serves as an estimate of the predictive uncertainty, reflecting the variance among the trees' predictions.

We used the RF implementation from scikit-learn version 1.2.2 in Python 3.10. For our active learning model, all hyperparameters were set to their default values in scikit-learn, except for the number of estimators (trees), which we increased to 150. This configuration provided us with 150 individual tree predictions, enabling us to calculate the standard deviation of these predictions for uncertainty quantification. 

\subsection{Deep Ensembles}\label{app:dehyp}

DEs are an ensemble learning technique where multiple neural networks are trained independently, each starting from different random initializations or using subsets of the data. By averaging the predictions of these independently trained models, deep ensembles can provide a natural way to estimate uncertainty \cite{abdar2021review}. DEs were utilized for the active learning procedure in the SBR benchmark problem. 

We constructed an ensemble of 10 neural network models, where each model is a pipeline that first scales the input features using $\texttt{MinMaxScaler}$ and then applies an MLP with hidden layers consisting of 100, 200, and 100 neurons. To introduce diversity among the models in the ensemble, each MLP was initialized with a different random seed. The MinMax scaler and the MLP implemented in our active learning code were from the scikit-learn 1.2.2 module. The MLP used default hyperparameters such as: the activation function was set to ReLU, the optimizer was Adam, and the learning rate was constant with an initial value of 0.001. The number of epochs was 200, with an L2 regularization parameter alpha set as 0.0001.

\section{Benchmark Models and Data}\label{app:benchmarks}

In this section, we provide comprehensive details on the construction of the inverse design benchmark datasets for all problems, as well as the training procedures and accuracy of the HF surrogates ($\mathbf{H}$). For each benchmark, the HF surrogates were used to evaluate the sampled design points (either with active learning or with samplers used for comparison) in order to form the TNN training datasets.

\subsection{Airfoil Inverse Design}\label{app:airfoil}

To address the airfoil inverse design (AID) problem, we construct a dataset to train the HF surrogate $\mathbf{H}$ by varying key parameters of NACA 4-digit airfoils and flow conditions. Specifically, we vary the maximum camber $m$, the position of maximum camber $p$, the maximum thickness $t$ of the airfoil, the Reynolds number $\mathrm{Re}$ of the fluid flow, and the angle of attack $\alpha$. Note that $m$, $p$, and $t$ are normalized by the chord length of the airfoil and thus take values between 0 and 1. The details on how to compute the geometrical coordinates of the NACA 4-digit airfoil shapes are given in the work by \cite{ladson1996computer}.

For each combination of these parameters, we obtain the pressure coefficient curve $\mathbf{C_p}$ by running simulations with XFOIL 6.99 software. XFOIL is a computational tool for the design and analysis of subsonic isolated airfoils, combining a panel method for potential flow with an integral boundary layer formulation.  Tab. \ref{tab:lbub_aid} shows the lower and upper boundaries of the design vector that contains the flow and shape parameters.

\begin{table}[h]
	\centering
	\caption{Lower and upper boundaries of the AID benchmark parameters used to generate the dataset.}
	\begin{tabular}{lcc}
		\hline
		\textbf{Parameter} & \textbf{Lower Bound} & \textbf{Upper Bound} \\
		\hline
		Maximum Camber, $m$ & 0.02 & 0.09 \\
		Position of Maximum Camber, $p$ & 0.2 & 0.7 \\
		Maximum Thickness, $t$ & 0.06 & 0.15 \\
		Reynolds Number, $\mathrm{Re}$ & $4 \times 10^6$ & $6 \times 10^6$ \\
		Angle of Attack, $\alpha$ & $0^\circ$ & $7^\circ$ \\
		\hline
	\end{tabular}
	\label{tab:lbub_aid}
\end{table}

Moreover, the total number of simulations that formed the initial dataset for the AID benchmark was 12,223. The dataset was divided into training, validation, and testing sets. Specifically, 70\% of the data was allocated to the training set, and the remaining 30\% was used as the testing set. Within the training set, 10\% was reserved as a validation set. Consequently, the final dataset comprised 63\% for training, 7\% for validation, and 30\% for testing. The Extreme Gradient Boosting (XGBoost) algorithm by \cite{chen2016xgboost} was used to train the HF surrogate ($\mathbf{H}$). The XGBoost algorithm was chosen as it excels in modeling structured tabular data \cite{shwartz2022tabular}. We used  the  xgboost 2.0.3 Python module with default hyperparameters, except for the ones adjusted as shown in  Tab. \ref{tab:xgbhyp_aid}. 

\begin{figure}[ht]
	\centering
	\begin{subfigure}[b]{0.49\textwidth}
		\includegraphics[width=\linewidth]{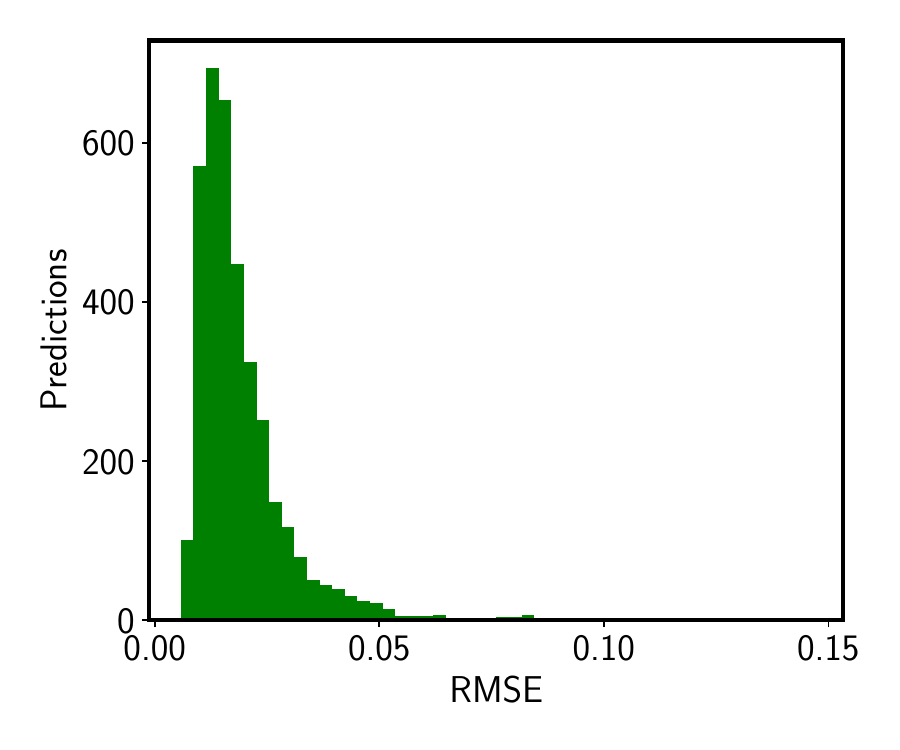}    \caption{}
		\label{fig:FigureA1a}
	\end{subfigure}
	\begin{subfigure}[b]{0.49\textwidth}
		\includegraphics[width=\linewidth]{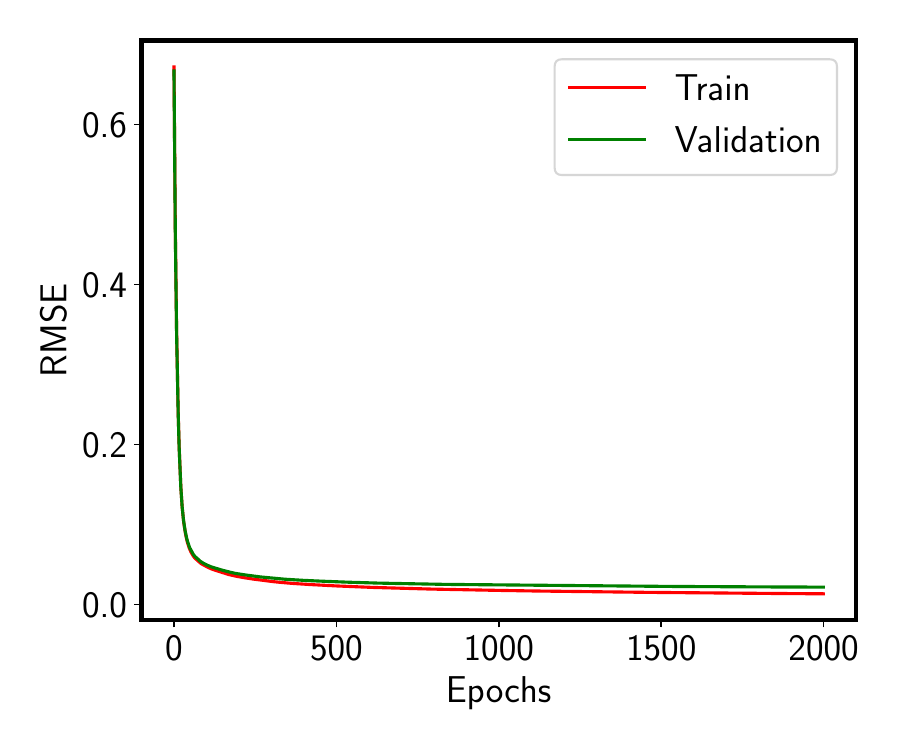}     
		\caption{}
		\label{fig:FigureA1b} 
	\end{subfigure}
	\begin{subfigure}[b]{0.49\textwidth}
		\includegraphics[width=\linewidth]{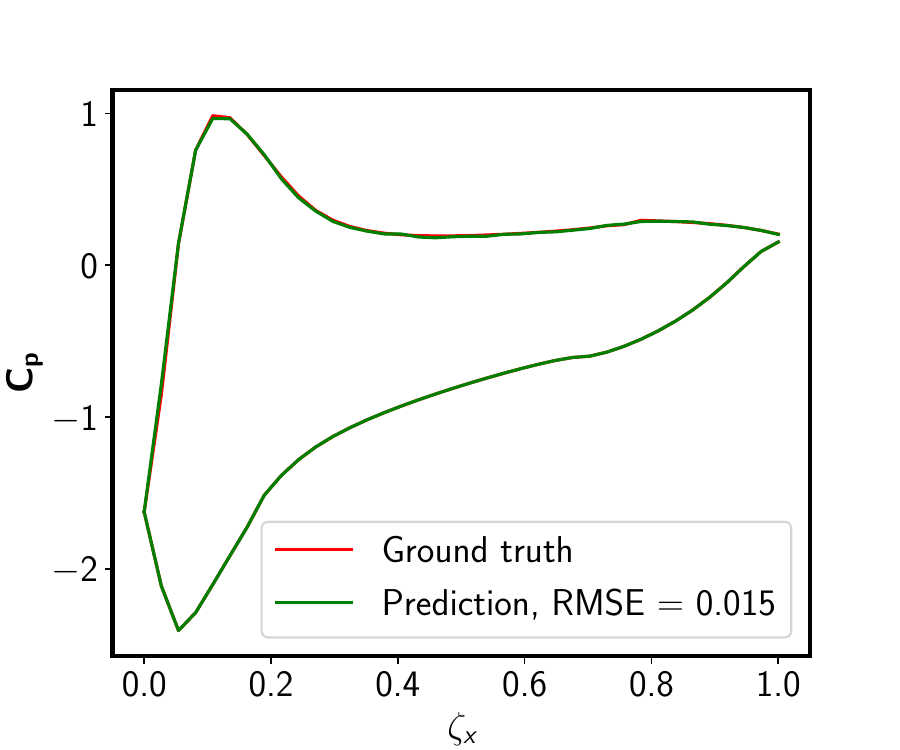}     
		\caption{}
		\label{fig:FigureA1c} 
	\end{subfigure}
	\caption[HF surrogate performance on AID benchmark]{Performance evaluation of the HF surrogate ($\mathbf{H}$) on the AID benchmark: (a) RMSE distribution between predicted and test set $\mathbf{C_p}$ curves (lower is better). (b) Training and validation loss curves for $\mathbf{H}$ using XGBoost algorithm. (c) Representative comparison between predicted and test set ground truth $\mathbf{C_p}$ curves with corresponding RMSE value.}
	\label{fig:FigureA1}
\end{figure}

Using the airfoil flow and shape parameters as inputs, the model was trained to predict the pressure coefficient curves, $\mathbf{C_p}$. Fig.~\ref{fig:FigureA1} illustrates the overall performance of the model on the testing dataset, which comprises 3,667 data instances. The model achieved an RMSE of 0.022 and an $R^2$ of 0.996, indicative of high predictive accuracy. Furthermore, Fig.~\ref{fig:FigureA1a} presents the distribution of RMSE values for each prediction compared to its corresponding test set curve. Fig.~\ref{fig:FigureA1b} displays the training and validation loss curves of the XGBoost model, demonstrating high stability and low error throughout the training process. Finally, Fig.~\ref{fig:FigureA1c} shows an example of a predicted $\mathbf{C_p}$ curve alongside the ground truth $\mathbf{C_p}$ curve from the test set, and it could be noticed that both curves are hardly distinguishable.

\begin{table}[h]
	\centering
	\caption{Hyperparameters of XGBoost algorithm used to train the AID HF surrogate.}
	\begin{tabular}{ll}
		\hline
		\textbf{Hyperparameter} & \textbf{Value} \\
		\hline
		Objective Function (\texttt{objective})            & \texttt{reg:squarederror} \\
		Evaluation Metric (\texttt{eval\_metric})          & \texttt{rmse} \\
		Learning Rate (\texttt{eta})                       & 0.1 \\
		Maximum Tree Depth (\texttt{max\_depth})           & 3 \\
		Subsample Ratio (\texttt{subsample})               & 0.8 \\
		Column Subsample Ratio (\texttt{colsample\_bytree}) & 0.8 \\
		L2 Regularization Term (\texttt{reg\_lambda})      & 10 \\
		Number of Estimators (\texttt{n\_estimators})      & 2000 \\
		Early Stopping Rounds (\texttt{early\_stopping\_rounds}) & 10 \\
		\hline
	\end{tabular}
	\label{tab:xgbhyp_aid}
\end{table}

\newpage
\subsection{Photonic Surface Inverse Design}\label{app:photonic}

To train the HF surrogate ($\mathbf{H}$) for the photonic surface inverse design problem, we utilized experimentally obtained data from our prior work \cite{grbcic2024inverse}. This data was generated by varying the laser manufacturing parameters—Laser power (L$_p$), Scanning speed (S$_s$), and Spacing of the textures (S$_p$)—to create photonic surfaces with Inconel alloy as the base material. The experimental setup is thoroughly described in our previous studies \cite{park2024inverse, grbcic2024inverse}. The total experimental dataset was comprised of 11,759 instances, 72\% of the instances were used for training, while 28\% were used for testing of the model. The lower and upper boundaries of these laser manufacturing parameters are provided in Table \ref{tab:lbub_psid}.

\begin{table}[h]
	\centering
	\caption{Lower and upper boundaries of the PSID benchmark parameters used to generate the dataset.}
	\begin{tabular}{lcc}
		\hline
		\textbf{Parameter} & \textbf{Lower Bound} & \textbf{Upper Bound} \\
		\hline
		Laser power, L$_p$ & 0.2 W & 1.3 W\\
		Scanning speed, S$_s$ & 10 mm/s & 700 mm/s \\
		Spacing, S$_p$ & 0.02 $\mu$m & 28 $\mu$m \\
		\hline
	\end{tabular}
	\label{tab:lbub_psid}
\end{table}

As outlined in our previous work \cite{grbcic2025artificial}, we employed the RF algorithm in combination with Principal Component Analysis (PCA) to train the HF surrogate ($\mathbf{H}$) for this benchmark. PCA was used to reduce the dimensionality of the 822-dimensional spectral emissivity curves to 10 principal components, thereby enhancing computational efficiency. The laser manufacturing parameters served as inputs to the model, while the PCA-compressed spectral emissivity curves $\mathbf{\epsilon}$ were the outputs. In the combined RF-PCA model, the RF predicted sets of 10 principal components, which were then inversely transformed using the PCA model to reconstruct the original spectral emissivity space.

We utilized the RF and PCA implementations from the scikit-learn 1.2.2 Python module \cite{pedregosa2011scikit}. All hyperparameters of the RF model were set to their default values except for the number of trees ($\texttt{n\_estimators}$), which was set to 450, and the maximum depth ($\texttt{max\_depth}$), which was set to 10. 

\begin{figure}[H]
	\centering
	\begin{subfigure}[t]{0.49\textwidth}
		\includegraphics[width=\linewidth]{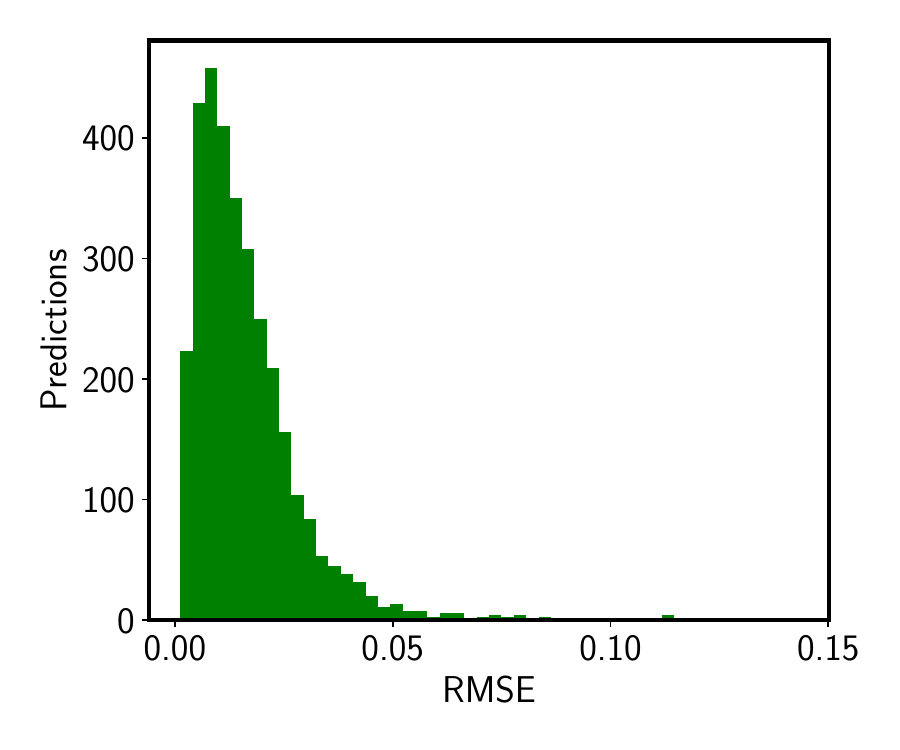}
		\caption{}
		\label{fig:FigureA2a}
	\end{subfigure}
	\begin{subfigure}[t]{0.49\textwidth}
		\includegraphics[height=5.74cm]{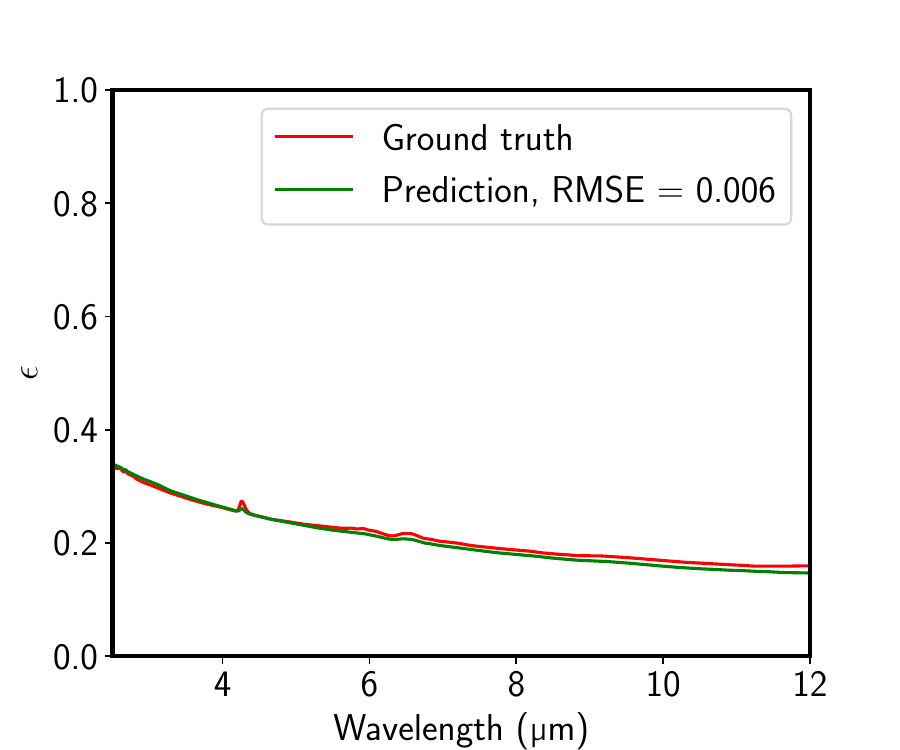}
		\caption{}
		\label{fig:FigureA2b}
	\end{subfigure}
	\caption[HF surrogate performance on PSID benchmark]{Performance evaluation of the HF surrogate ($\mathbf{H}$) on the PSID benchmark: (a) RMSE distribution between predicted and test set $\mathbf{\epsilon}$ curves (lower is better). (b) Representative comparison between predicted and test set $\epsilon$ curves with corresponding RMSE value.}
	\label{fig:FigureA2}
\end{figure}

The RF-PCA model achieved an RMSE of 0.0212 and an R$^2$ of 0.977, indicating high predictive accuracy. Fig. \ref{fig:FigureA2} illustrates the performance of the RF-PCA model in detail. Specifically, Fig.  \ref{fig:FigureA2a} shows the distribution of RMSE values for all predicted spectral emissivity curves ($\mathbf{\epsilon}$) when compared to the experimental test set curves. Fig.  \ref{fig:FigureA2b} presents a juxtaposition of a predicted emissivity curve and its corresponding experimental ground truth curve, with the RMSE value indicated in the legend.

\subsection{Scalar Boundary Reconstruction}\label{app:scalar}

To generate the dataset for the HF surrogate  $\mathbf{H}$, we employed the open-source computational fluid dynamics (CFD) toolbox OpenFOAM version 9 \cite{jasak2007openfoam}. OpenFOAM utilizes the finite volume method to numerically solve partial differential equations (PDEs) that arise in fluid dynamics and heat transfer. Specifically, we used the $\texttt{laplacianFoam}$ solver provided by OpenFOAM, which is designed to solve the unsteady diffusion equation. This solver allowed us to simulate the temporal and spatial evolution of the diffusion process with high fidelity, providing accurate data for training and validating our model. The unsteady diffusion equation is defined as:

\begin{equation}
	\frac{\partial c}{\partial t} = D \nabla^2 c \quad \text{in } \Omega, \quad t \in [0, t_{max}]
	\label{eqn:diffusion}
\end{equation}

where $c$ is the dimensionless scalar value in the domain, $D$ is the coefficient of diffusion (defined as 1 m$^2$/s), $t$ is the time (s), and $t_{max}$ denotes the end time of the simulation, and $\Omega$ is the 2D domain. The $t_{max}$ is defined as 0.1 s and is treated as a converged state. The scalar boundary values are a Dirichlet boundary condition type along the top of the domain: 

\begin{equation}
	\begin{aligned}
		c = g(\mathbf{c_{BC}}, t) \quad \text{on } \partial \Omega_{top}, \quad t \in [0, t_{max}].
		\label{eqn:diffusion_bc_top}
	\end{aligned}
\end{equation}

The remaining parts of the 2D domain $\partial \Omega_{other}$ (left, right, bottom) are the Neumann boundary condition type:

\begin{equation}
	\begin{aligned}
		\frac{\partial c}{\partial \mathbf{n}} = 0 \quad \text{on } \partial \Omega_{other}, \quad t \in [0, t_{max}],
		\label{eqn:diffusion_bc_other}
	\end{aligned}
\end{equation}

where $\mathbf{n}$ is the unit normal vector pointing outward from the domain. The finite volume mesh used in the simulations consisted of a total of 400 cells, providing adequate spatial resolution for capturing the diffusion process. Specifically, the top boundary of the domain was discretized into 20 cells, corresponding to the 20 elements of the boundary condition array $\mathbf{c_{BC}}$.

For each simulation, we randomly varied the values of the boundary condition array $\mathbf{c_{BC}}$ (along the top $\partial \Omega_{top}$, ensuring that each component was within the range of 0 to 30, (for $\mathbf{c_{BC}}$ $\in$ $\mathbb{R}^{20}$). This randomization introduced a diverse set of boundary conditions, allowing the model to learn from a wide variety of scenarios and improving its generalization capabilities. Upon completing each simulation, we obtained scalar values from scattered measurement locations within the two-dimensional domain, as illustrated in Fig. \ref{fig:Figure2c}. These collected values were compiled into the scalar measurements vector $\mathbf{c}$. Additional details on the construction of the random $\mathbf{c_{BC}}$  generator and the coordinates of the 2D scattered measurement locations are provided in our previous work (\cite{grbcic2024efficient}).

The generated dataset consisted of 10,000 instances. We allocated 70\% of the data to the training set and the remaining 30\% to the testing set. Within the training set, 10\% was reserved for validation purposes. Consequently, the final dataset comprised 63\% for training, 7\% for validation, and 30\% for testing. We used the XGBoost algorithm to train the HF surrogate ($\mathbf{H}$) with the hyperparameters specified in Tab. \ref{tab:xgbhyp_sbr}. All other hyperparameters were set to their default values as provided by the xgboost 2.0.3 Python module.

\begin{figure}[H]
	\centering
	\begin{subfigure}[b]{0.49\textwidth}
		\includegraphics[width=\linewidth]{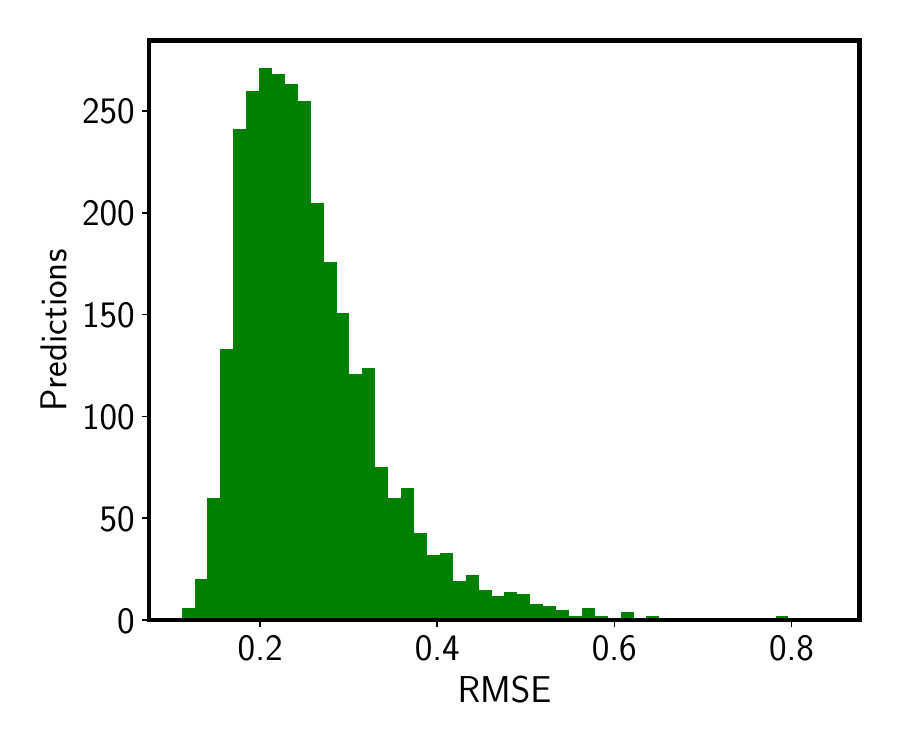}    \caption{}
		\label{fig:FigureA3a}
	\end{subfigure}
	\begin{subfigure}[b]{0.49\textwidth}
		\includegraphics[width=\linewidth]{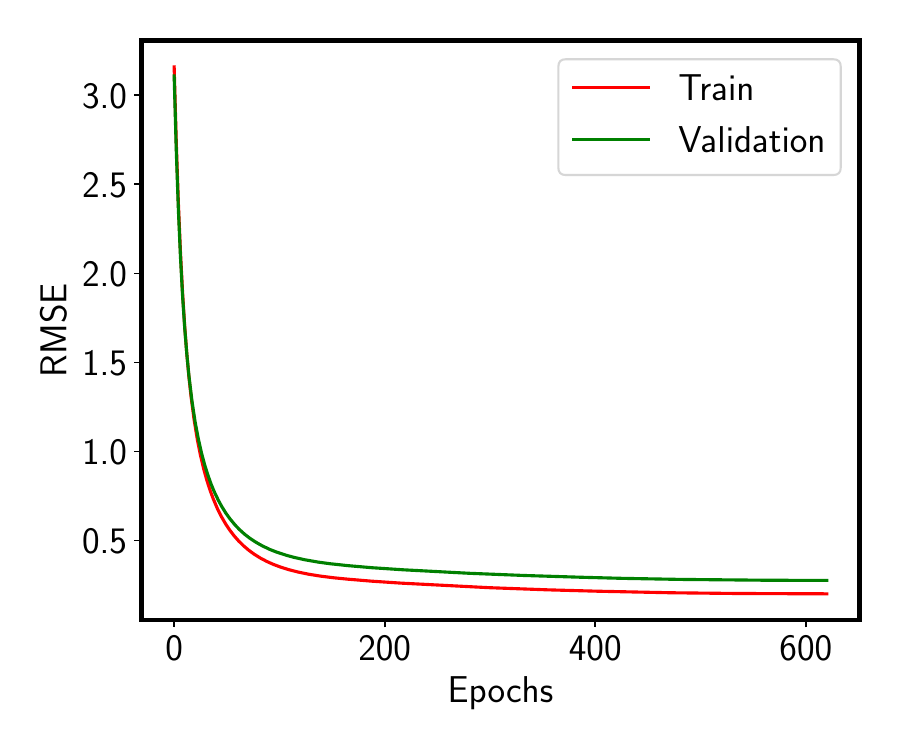}     
		\caption{}
		\label{fig:FigureA3b} 
	\end{subfigure}
	\begin{subfigure}[b]{0.49\textwidth}
	\includegraphics[width=\linewidth]{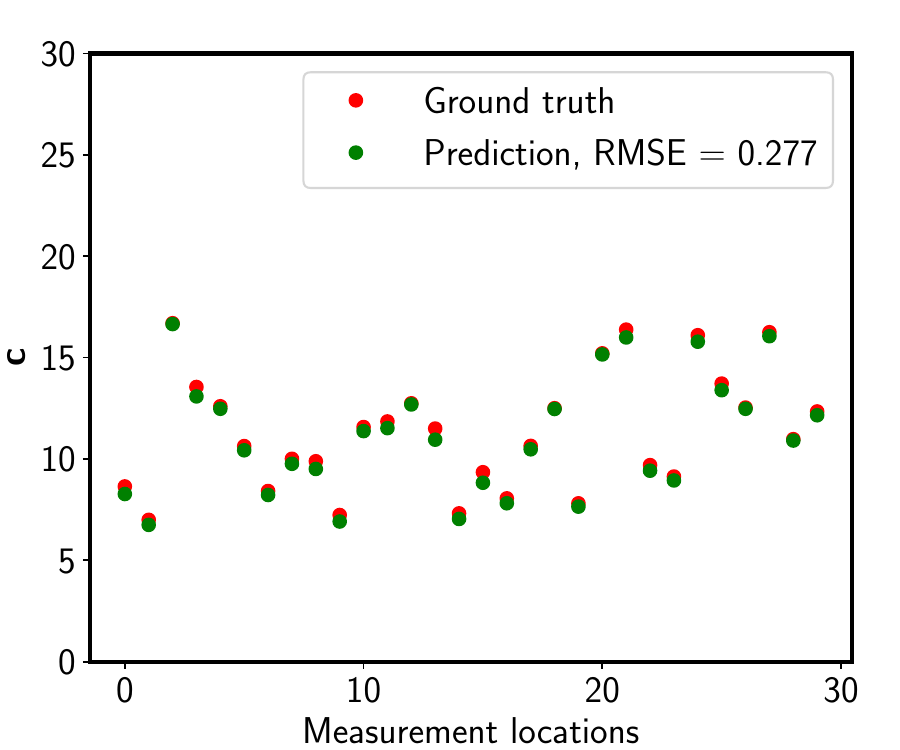}     
	\caption{}
	\label{fig:FigureA3c} 
\end{subfigure}
	\caption[HF surrogate performance on SBR benchmark]{Performance evaluation of the HF surrogate ($\mathbf{H}$) on the SBR benchmark: (a) RMSE distribution between predicted and test set $\mathbf{c}$ measured values (lower is better). (b) Training and validation loss curves for $\mathbf{H}$ using XGBoost algorithm. (c) Representative comparison between predicted and test set $\mathbf{c}$ values with corresponding RMSE value.}
	\label{fig:FigureA3}
\end{figure}

The XGBoost model achieved an overall RMSE of 0.270 and an R$^2$ of 0.984, indicating high performance. Fig. \ref{fig:FigureA3} shows the performance of the model. Specifically, Fig. \ref{fig:FigureA3a} presents the RMSE distribution of each predicted set of measurements ($\mathbf{c}$) when compared to the test set values. Furthermore, Fig. \ref{fig:FigureA3b} displays the training and validation curves of the XGBoost model, and Fig. \ref{fig:FigureA3c} shows an example of the test set ground truth values and the predicted set of measurements, with the RMSE value provided in the legend.

\begin{table}[h]
	\centering
	\caption{Hyperparameters of XGBoost algorithm used to train the SBR HF surrogate.}
	\begin{tabular}{ll}
		\hline
		\textbf{Hyperparameter} & \textbf{Value} \\
		\hline
		Objective Function (\texttt{objective})            & \texttt{reg:squarederror} \\
		Evaluation Metric (\texttt{eval\_metric})          & \texttt{rmse} \\
		Learning Rate (\texttt{eta})                       & 0.1 \\
		Maximum Tree Depth (\texttt{max\_depth})           & 3 \\
		L1 Regularization Term (\texttt{reg\_alpha})      & 20 \\
		L2 Regularization Term (\texttt{reg\_lambda})      & 50 \\
		Number of Estimators (\texttt{n\_estimators})      & 3000 \\
		Early Stopping Rounds (\texttt{early\_stopping\_rounds}) & 5 \\
		\hline
	\end{tabular}
	\label{tab:xgbhyp_sbr}
\end{table}

\section{Forward Model Benchmark Results}\label{app:forward_model}

In this section, we present the results of the forward model $\mathbf{M}$ for all three inverse design benchmarks. Specifically, we compare the performance of $\mathbf{M}$ when trained using active learning with its performance when trained using other sampling methods (Random, Latin Hybercube sampling, Best Candidate sampling, GreedyFP sampling). The comparison results are presented through box plots and statistical summaries.

\subsection{Forward Model Airfoil Inverse Design Results}
\label{app:forward_model_aid}

\begin{figure}[H]
\centering
\begin{subfigure}[b]{0.49\textwidth}
   \includegraphics[width=\linewidth]{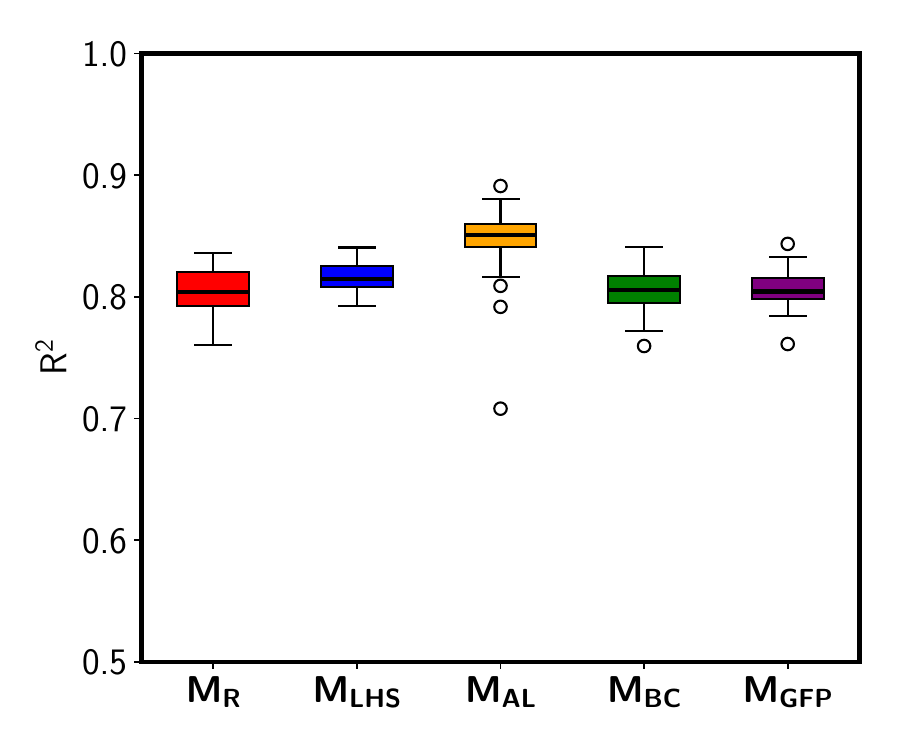}    \caption{}
   \label{fig:FigureC1a}
\end{subfigure}
\begin{subfigure}[b]{0.49\textwidth}
   \includegraphics[width=\linewidth]{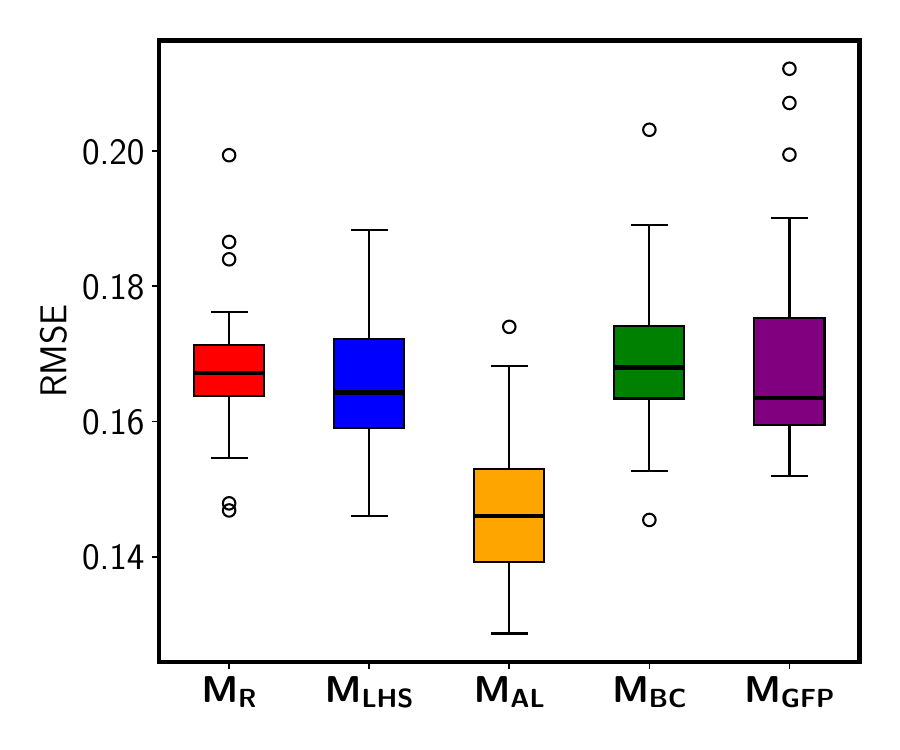}     
   \caption{}
   \label{fig:FigureC1b} 
\end{subfigure}
\begin{subfigure}[b]{0.49\textwidth}
   \includegraphics[width=\linewidth]{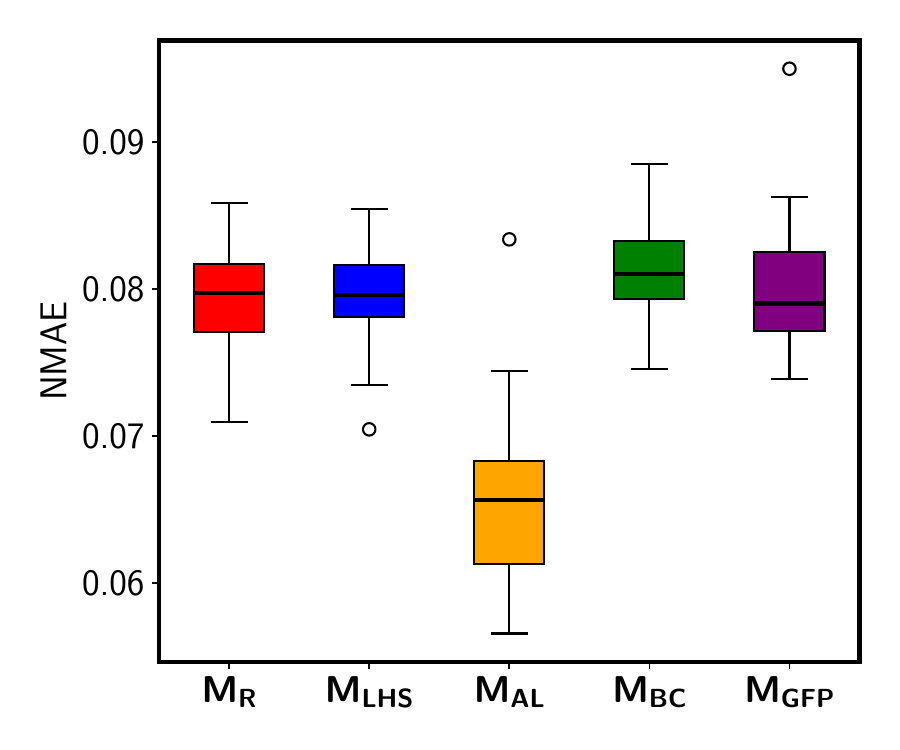}
   \caption{}
   \label{fig:FigureC1c}
\end{subfigure}
\caption[]{Forward model ($\mathbf{M}$) performance on AID benchmark problem using different dataset generation methods: (a) R$^2$ (higher is better), (b) RMSE (lower is better), and (c) NMAE (lower is better). Subscripts in $\mathbf{M_{R}}$ denote the sampling method (e.g., $\mathbf{M_{R}}$ for random sampling).}
\label{fig:FigureC1}
\end{figure}

In Fig.\ \ref{fig:FigureC1} and Tab.\ \ref{tab:aid_m_summary}, we present the results for the AID benchmark. Across all three performance metrics—R$^2$, RMSE, and NMAE—the active learning approach ($\mathbf{M_{AL}}$) consistently outperforms the other sampling methods. Specifically, Fig.\ \ref{fig:FigureC1a} displays the R$^2$ score, where $\mathbf{M_{AL}}$ achieves the highest value of R$^2$=0.84, as reported in Tab.\ \ref{tab:aid_m_summary}. Similarly, Fig.\ \ref{fig:FigureC1b} shows the RMSE results, with $\mathbf{M_{AL}}$ attaining the lowest RMSE of 0.147. Finally, Fig.\ \ref{fig:FigureC1c} illustrates the NMAE scores, where $\mathbf{M_{AL}}$ again achieves the lowest value of NMAE=0.065. These results, summarized in Tab.\ \ref{tab:aid_m_summary}, confirm the superior performance of the active learning approach over the other samplers in the AID benchmark.

\begin{table}[ht]
	\centering
	\caption{Statistical analysis of the forward model ($\mathbf{M}$) performance on AID benchmark problem using different dataset generation methods. Bold values indicate best performance in each metric. RMSE and NMAE values should be as low as possible, while R$^2$ should be as high as possible.}
	\begin{tabular}{lccccc}
		\hline
		\textbf{Metric} & \textbf{Method} & \textbf{Mean} & \textbf{Std} & \textbf{Max} & \textbf{Min} \\
		\hline
RMSE & $\mathbf{M_{R}}$ & 0.1679 & 0.0105 & 0.1994 & 0.1469 \\
RMSE & $\mathbf{M_{LHS}}$ & 0.1658 & 0.0087 & 0.1884 & 0.1461 \\
RMSE & $\mathbf{M_{AL}}$ & \textbf{0.1470} & 0.0115 & \textbf{0.1740} & \textbf{0.1287} \\
RMSE & $\mathbf{M_{BC}}$ & 0.1691 & 0.0112 & 0.2031 & 0.1455 \\
RMSE & $\mathbf{M_{GFP}}$ & 0.1696 & 0.0155 & 0.2122 & 0.1520 \\
\hline
R$^2$ & $\mathbf{M_{R}}$ & 0.8040 & 0.0197 & 0.8360 & 0.7601 \\
R$^2$ & $\mathbf{M_{LHS}}$ & 0.8158 & 0.0120 & 0.8405 & \textbf{0.7921} \\
R$^2$ & $\mathbf{M_{AL}}$ & \textbf{0.8444} & 0.0330 & \textbf{0.8911} & 0.7080 \\
R$^2$ & $\mathbf{M_{BC}}$ & 0.8052 & 0.0188 & 0.8406 & 0.7595 \\
R$^2$ & $\mathbf{M_{GFP}}$ & 0.8060 & 0.0160 & 0.8435 & 0.7612 \\
\hline
NMAE & $\mathbf{M_{R}}$ & 0.0796 & 0.0032 & 0.0858 & 0.0710 \\
NMAE & $\mathbf{M_{LHS}}$ & 0.0798 & 0.0033 & 0.0854 & 0.0704 \\
NMAE & $\mathbf{M_{AL}}$ & \textbf{0.0655} & 0.0059 & \textbf{0.0834} & \textbf{0.0565} \\
NMAE & $\mathbf{M_{BC}}$ & 0.0814 & 0.0035 & 0.0885 & 0.0746 \\
NMAE & $\mathbf{M_{GFP}}$ & 0.0801 & 0.0045 & 0.0950 & 0.0738 \\
		\hline
	\end{tabular}
\label{tab:aid_m_summary}
\end{table}

\newpage
\subsection{Forward Model Photonic Surface Inverse Design Results}
\label{app:forward_model_psid}

\begin{figure}[H]
	\centering
	\begin{subfigure}[b]{0.49\textwidth}
		\includegraphics[width=\linewidth]{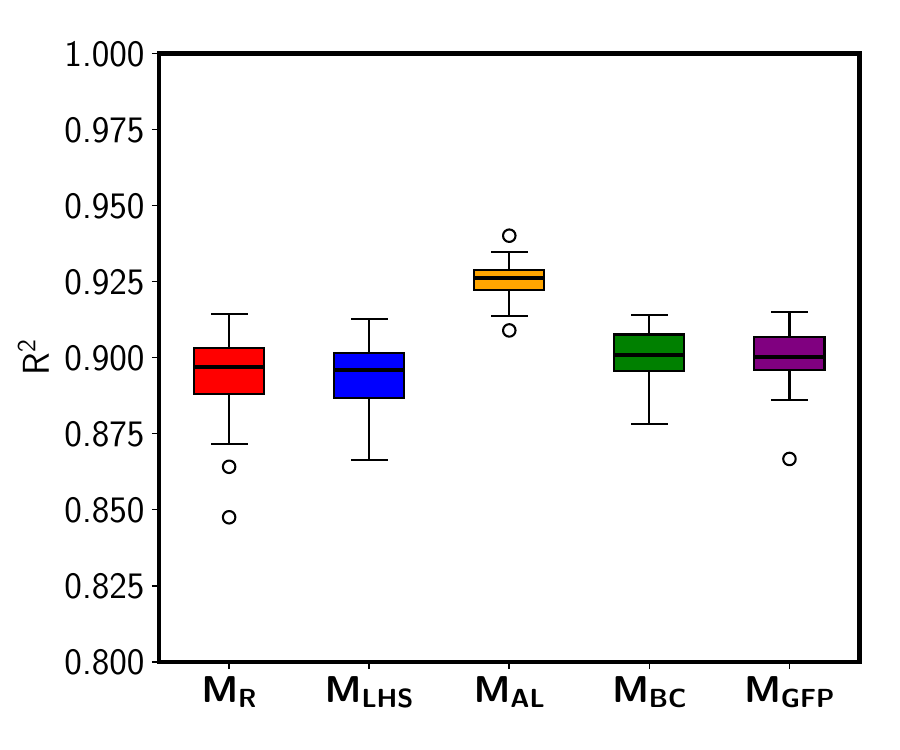}    \caption{}
		\label{fig:FigureC2a}
	\end{subfigure}
	\begin{subfigure}[b]{0.49\textwidth}
		\includegraphics[width=\linewidth]{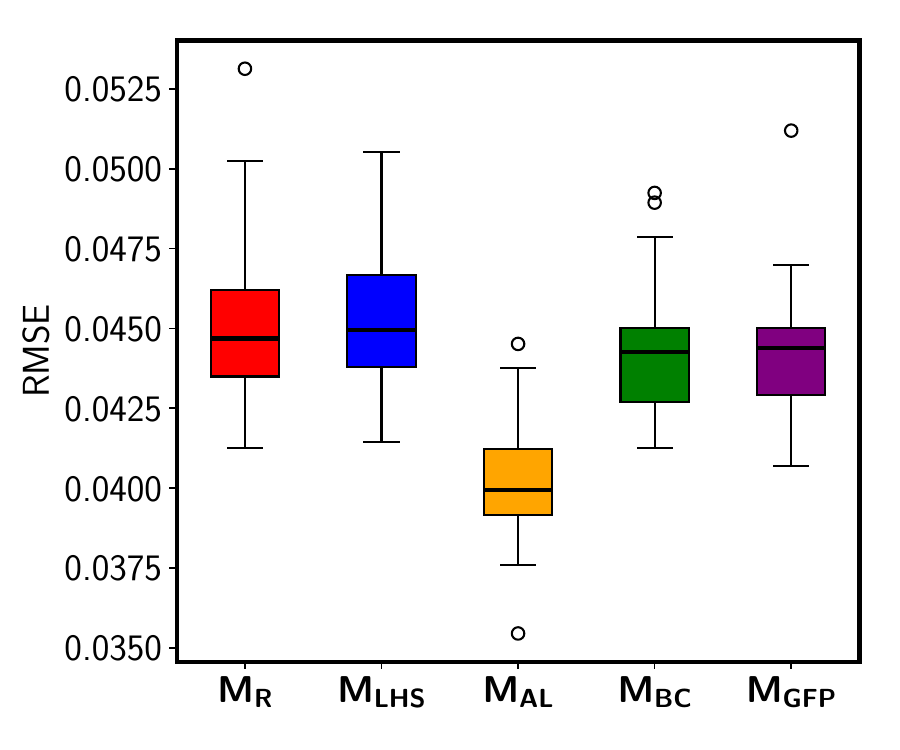}     
		\caption{}
		\label{fig:FigureC2b} 
	\end{subfigure}
	\begin{subfigure}[b]{0.49\textwidth}
		\includegraphics[width=\linewidth]{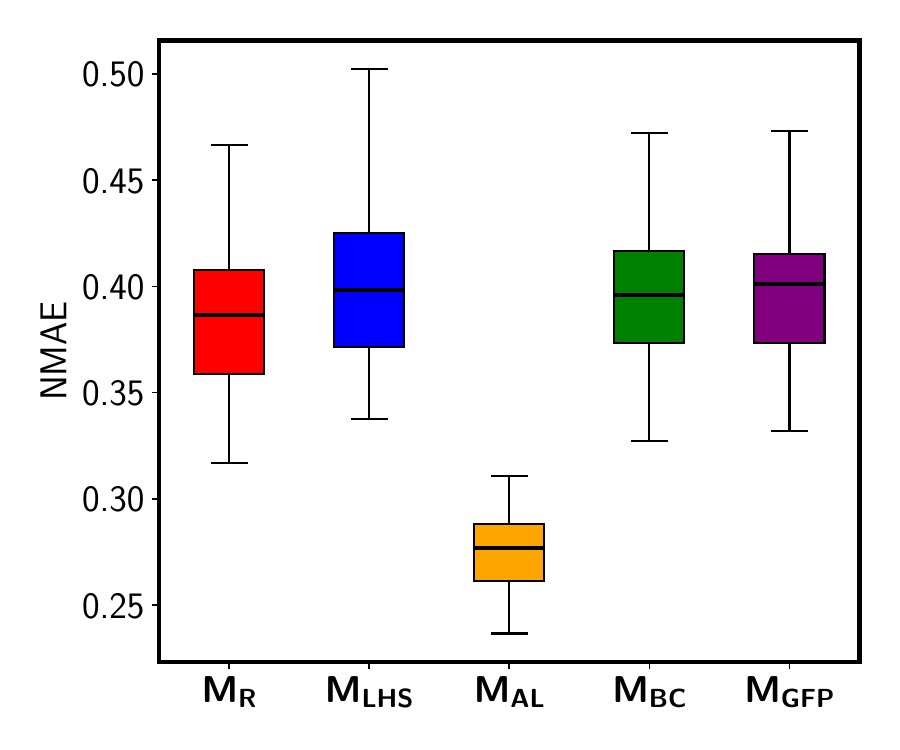}
		\caption{}
		\label{fig:FigureC2c}
	\end{subfigure}
\caption[]{Forward model ($\mathbf{M}$) performance on PSID benchmark problem using different dataset generation methods: (a) R$^2$ (higher is better), (b) RMSE (lower is better), and (c) NMAE (lower is better). Subscripts in $\mathbf{M_{R}}$ denote the sampling method (e.g., $\mathbf{M_{R}}$ for random sampling).}
	\label{fig:FigureC2}
\end{figure}

In Fig.\ \ref{fig:FigureC2} and Tab.\ \ref{tab:psid_m_summary}, we present the results for the PSID benchmark. Across all three performance metrics—R$^2$, RMSE, and NMAE—the active learning approach ($\mathbf{M_{AL}}$) consistently outperforms the other sampling methods. Specifically, Fig.\ \ref{fig:FigureC2a} displays the R$^2$ score, where $\mathbf{M_{AL}}$ achieves the highest value of R$^2$=0.92, as reported in Tab.\ \ref{tab:psid_m_summary}. Similarly, Fig.\ \ref{fig:FigureC2b} shows the RMSE results, with $\mathbf{M_{AL}}$ attaining the lowest RMSE of 0.04. Finally, Fig.\ \ref{fig:FigureC2c} illustrates the NMAE scores, where $\mathbf{M_{AL}}$ again achieves the lowest value of NMAE=0.275. These results, summarized in Tab.\ \ref{tab:psid_m_summary}, confirm the superior performance of the active learning approach over the other samplers in the PSID benchmark.

\begin{table}[ht]
	\centering
	\caption{Statistical analysis of the forward model ($\mathbf{M}$) performance on PSID benchmark problem using different dataset generation methods. Bold values indicate best performance in each metric. RMSE and NMAE values should be as low as possible, while R$^2$ should be as high as possible.}
	\begin{tabular}{lccccc}
		\hline
		\textbf{Metric} & \textbf{Method} & \textbf{Mean} & \textbf{Std} & \textbf{Max} & \textbf{Min} \\
		\hline
RMSE & $\mathbf{M_{R}}$ & 0.0451 & 0.0026 & 0.0531 & 0.0413 \\
RMSE & $\mathbf{M_{LHS}}$ & 0.0454 & 0.0021 & 0.0505 & 0.0414 \\
RMSE & $\mathbf{M_{AL}}$ & \textbf{0.0402} & 0.0020 & \textbf{0.0445} & \textbf{0.0354} \\
RMSE & $\mathbf{M_{BC}}$ & 0.0444 & 0.0021 & 0.0492 & 0.0413 \\
RMSE & $\mathbf{M_{GFP}}$ & 0.0443 & 0.0020 & 0.0512 & 0.0407 \\
\hline
R$^2$ & $\mathbf{M_{R}}$ & 0.8944 & 0.0147 & 0.9144 & 0.8475 \\
R$^2$ & $\mathbf{M_{LHS}}$ & 0.8937 & 0.0107 & 0.9128 & 0.8664 \\
R$^2$ & $\mathbf{M_{AL}}$ & \textbf{0.9251} & 0.0067 & \textbf{0.9401} & \textbf{0.9089} \\
R$^2$ & $\mathbf{M_{BC}}$ & 0.8996 & 0.0102 & 0.9140 & 0.8781 \\
R$^2$ & $\mathbf{M_{GFP}}$ & 0.8998 & 0.0097 & 0.9151 & 0.8667 \\
\hline
NMAE & $\mathbf{M_{R}}$ & 0.3856 & 0.0332 & 0.4667 & 0.3167 \\
NMAE & $\mathbf{M_{LHS}}$ & 0.4014 & 0.0384 & 0.5024 & 0.3376 \\
NMAE & $\mathbf{M_{AL}}$ & \textbf{0.2758} & 0.0172 & \textbf{0.3110} & \textbf{0.2367} \\
NMAE & $\mathbf{M_{BC}}$ & 0.3973 & 0.0341 & 0.4723 & 0.3273 \\
NMAE & $\mathbf{M_{GFP}}$ & 0.3961 & 0.0356 & 0.4729 & 0.3320 \\
		\hline
	\end{tabular}
\label{tab:psid_m_summary}
\end{table}

\newpage
\subsection{Forward Model Scalar Field Reconstruction Results}
\label{app:forward_model_sbr}

\begin{figure}[!h]
	\centering
	\begin{subfigure}[b]{0.49\textwidth}
		\includegraphics[width=\linewidth]{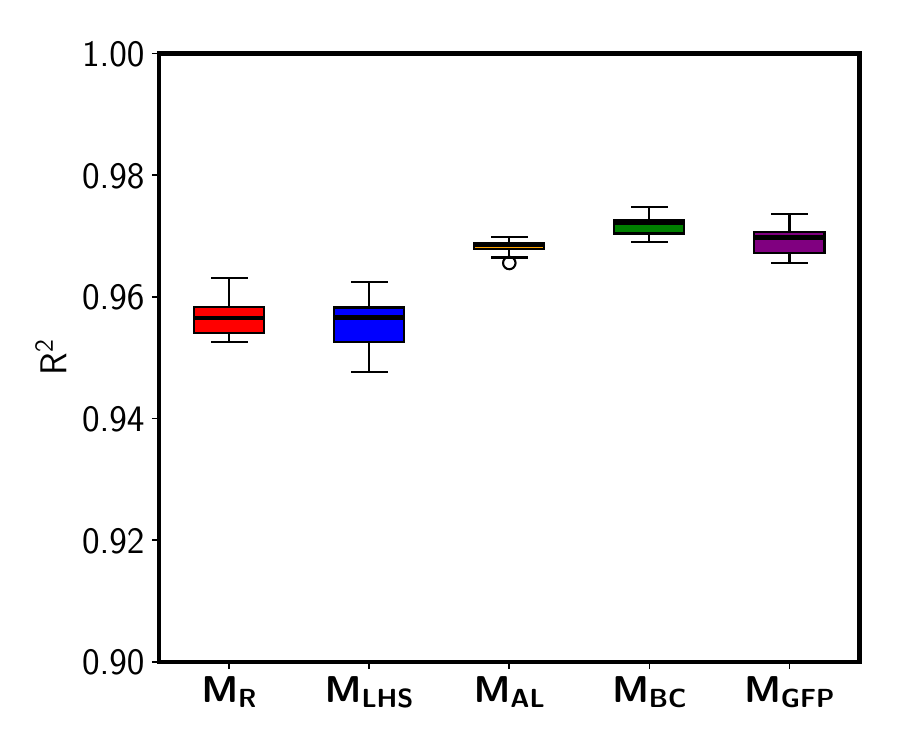}    \caption{}
		\label{fig:FigureC3a}
	\end{subfigure}
	\begin{subfigure}[b]{0.49\textwidth}
		\includegraphics[width=\linewidth]{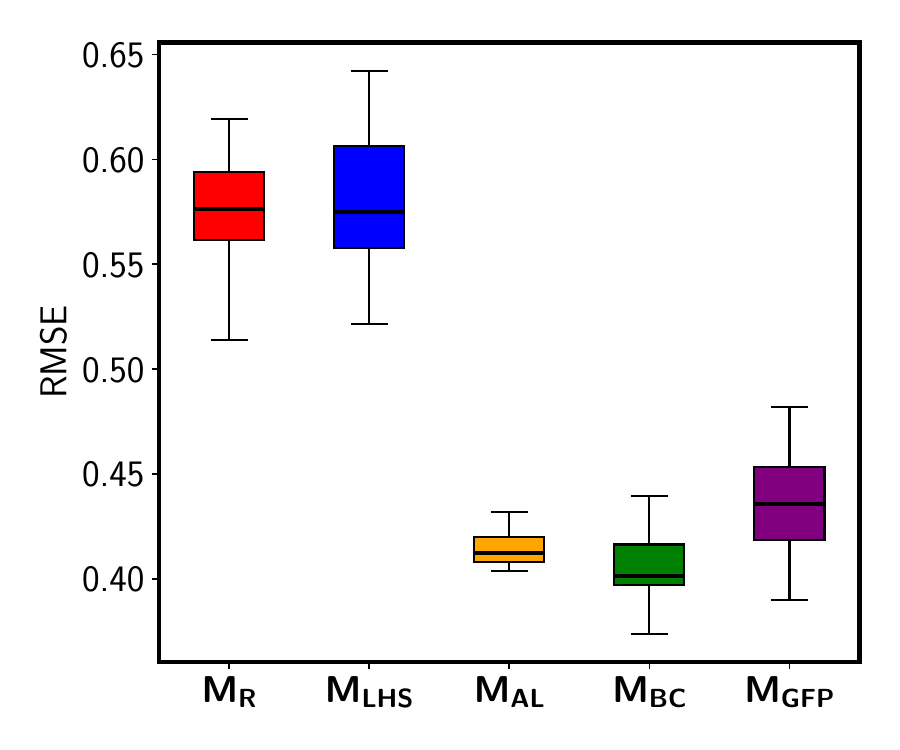}     
		\caption{}
		\label{fig:FigureC3b} 
	\end{subfigure}
	\begin{subfigure}[b]{0.49\textwidth}
		\includegraphics[width=\linewidth]{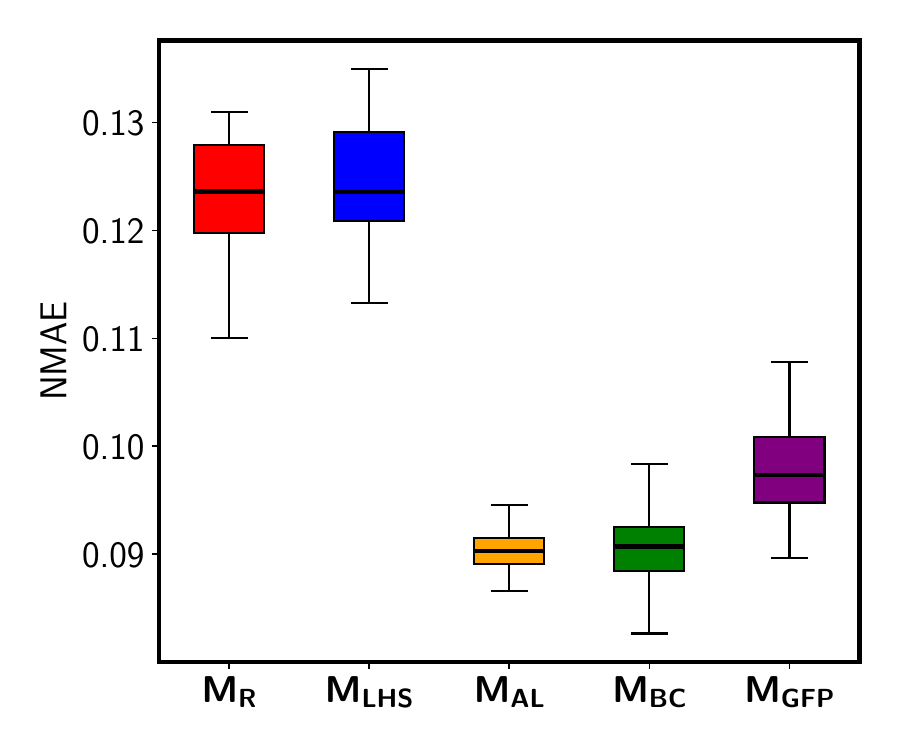}
		\caption{}
		\label{fig:FigureC3c}
	\end{subfigure}
\caption[]{Forward model ($\mathbf{M}$) performance on SBR benchmark problem using different dataset generation methods: (a) R$^2$ (higher is better), (b) RMSE (lower is better), and (c) NMAE (lower is better). Subscripts in $\mathbf{M_{R}}$ denote the sampling method (e.g., $\mathbf{M_{R}}$ for random sampling).}
	\label{fig:FigureC3}
\end{figure}

In Fig.\ \ref{fig:FigureC3} and Tab.\ \ref{tab:sbr_m_summary}, we present the results for the SBR benchmark. The performance of $\mathbf{M_{AL}}$ differs slightly from that in the AID and PSID benchmarks. In the SBR benchmark, the best-performing algorithm for the forward model is $\mathbf{M_{BC}}$; however, $\mathbf{M_{AL}}$ comes as a close second in the R$^2$ and RMSE metrics. Specifically, Fig.\ \ref{fig:FigureC3a} displays the R$^2$ scores, where $\mathbf{M_{BC}}$ achieves the highest value of R$^2$=0.97, as reported in Tab.\ \ref{tab:sbr_m_summary}, while $\mathbf{M_{AL}}$ obtains a score of 0.96. Similarly, Fig.\ \ref{fig:FigureC3b} shows the RMSE results, with $\mathbf{M_{BC}}$ attaining the lowest RMSE of 0.40, while $\mathbf{M_{AL}}$ achieves an RMSE of 0.41. However, Fig.\ \ref{fig:FigureC3c} illustrates the NMAE scores, where $\mathbf{M_{AL}}$ and $\mathbf{M_{BC}}$ both achieve the lowest value of NMAE=0.09. These results, summarized in Tab.\ \ref{tab:sbr_m_summary}, demonstrate that $\mathbf{M_{AL}}$ can also perform competitively.
\newpage

\newpage
\begin{table}[ht]
	\centering
	\caption{Statistical analysis of the forward model ($\mathbf{M}$) performance on SBR benchmark problem using different dataset generation methods. Bold values indicate best performance in each metric. RMSE and NMAE values should be as low as possible, while R$^2$ should be as high as possible.}
	\begin{tabular}{lccccc}
		\hline
		\textbf{Metric} & \textbf{Method} & \textbf{Mean} & \textbf{Std} & \textbf{Max} & \textbf{Min} \\
		\hline
RMSE & $\mathbf{M_{R}}$ & 0.5741 & 0.0259 & 0.6194 & 0.5138 \\
RMSE & $\mathbf{M_{LHS}}$ & 0.5793 & 0.0321 & 0.6423 & 0.5217 \\
RMSE & $\mathbf{M_{AL}}$ & 0.4143 & 0.0069 & \textbf{0.4317} & 0.4035 \\
RMSE & $\mathbf{M_{BC}}$ & \textbf{0.4047} & 0.0158 & 0.4393 & \textbf{0.3739} \\
RMSE & $\mathbf{M_{GFP}}$ & 0.4358 & 0.0239 & 0.4817 & 0.3898 \\
\hline
R$^2$ & $\mathbf{M_{R}}$ & 0.9565 & 0.0028 & 0.9631 & 0.9525 \\
R$^2$ & $\mathbf{M_{LHS}}$ & 0.9555 & 0.0040 & 0.9624 & 0.9477 \\
R$^2$ & $\mathbf{M_{AL}}$ & 0.9684 & 0.0009 & 0.9698 & 0.9656 \\
R$^2$ & $\mathbf{M_{BC}}$ & \textbf{0.9718} & 0.0016 & \textbf{0.9748} & \textbf{0.9689} \\
R$^2$ & $\mathbf{M_{GFP}}$ & 0.9695 & 0.0024 & 0.9736 & 0.9656 \\
\hline
NMAE & $\mathbf{M_{R}}$ & 0.1230 & 0.0053 & 0.1309 & 0.1100 \\
NMAE & $\mathbf{M_{LHS}}$ & 0.1245 & 0.0057 & 0.1350 & 0.1133 \\
NMAE & $\mathbf{M_{AL}}$ & \textbf{0.0903} & 0.0020 & \textbf{0.0946} & 0.0866 \\
NMAE & $\mathbf{M_{BC}}$ & 0.0906 & 0.0035 & 0.0984 & \textbf{0.0827} \\
NMAE & $\mathbf{M_{GFP}}$ & 0.0981 & 0.0046 & 0.1078 & 0.0897 \\
		\hline
	\end{tabular}
\label{tab:sbr_m_summary}
\end{table}

\end{appendices}

\bibliography{sn-bibliography}

\end{document}